\documentclass[11pt,letterpaper]{article}
\usepackage[left=0.88in,right=0.88in,top=0.80in,bottom=0.74in,footskip=24pt]{geometry}
\usepackage[T1]{fontenc}
\usepackage{newtxtext}
\usepackage{amsmath,amssymb,amsthm,mathtools}
\usepackage{newtxmath}
\usepackage{bm,mathrsfs}
\DeclareFontFamily{U}{rsfs}{}
\DeclareFontShape{U}{rsfs}{m}{n}{<-6>rsfs5<6-8>rsfs7<8->rsfs10}{}
\usepackage{booktabs,tabularx,array,longtable}
\usepackage{microtype}
\usepackage{tikz}
\usetikzlibrary{arrows.meta,calc}
\usepackage{placeins,float,needspace}
\usepackage{enumitem}
\usepackage{titlesec}
\usepackage[font=small,labelfont=bf,labelsep=period]{caption}
\usepackage[numbers,sort&compress]{natbib}
\usepackage[hidelinks,breaklinks=true]{hyperref}
\hypersetup{pdftitle={Dynamics of Hidden Learning Structure},pdfauthor={Qinyou Wang},pdfsubject={Hidden learning structure, regulation, acquisition, and coupled dynamics}}
\numberwithin{equation}{section}
\titleformat{\section}{\normalfont\fontsize{12}{14}\selectfont\bfseries}{\thesection}{0.75em}{}
\titleformat{\subsection}{\normalfont\normalsize\bfseries}{\thesubsection}{0.75em}{}
\titlespacing*{\section}{0pt}{2.0ex plus .4ex minus .2ex}{0.8ex plus .2ex}
\titlespacing*{\subsection}{0pt}{1.6ex plus .3ex minus .2ex}{0.6ex plus .2ex}
\titlespacing*{\paragraph}{0pt}{1.4ex plus .2ex}{0.75em}
\makeatletter
\renewcommand{\@maketitle}{%
  \begin{center}
    {\fontsize{19}{22}\selectfont\bfseries\@title\par}
    \vskip 12pt
    {\normalsize\@author\par}
  \end{center}\vskip 9pt}
\makeatother
\setlist{nosep,leftmargin=*}
\newtheorem{theorem}{Theorem}[section]
\newtheorem{proposition}[theorem]{Proposition}

\theoremstyle{definition}\newtheorem{definition}[theorem]{Definition}
\theoremstyle{remark}

\newcommand{\R}{\mathbb R}\newcommand{\E}{\mathbb E}

\newcommand{\CE}{\mathrm{CE}}\newcommand{\KL}{\mathrm{KL}}

\newcommand{\norm}[1]{\left\lVert #1\right\rVert}
\newcommand{\ip}[2]{\left\langle #1,#2\right\rangle}
\DeclareMathOperator{\Ran}{Ran}\DeclareMathOperator{\rank}{rank}
\DeclareMathOperator{\diag}{diag}\DeclareMathOperator{\tr}{tr}
\DeclareMathOperator{\softmax}{softmax}\DeclareMathOperator{\dom}{dom}

\title{Dynamics of Hidden Learning Structure}
\author{Qinyou Wang}
\date{}
\begin{document}
\maketitle
\begin{abstract}
A learner's present predictions do not fully determine how it will learn next. We study the dynamics of hidden learning structure through two related operations. Fiber Regulation Theory (FRT) describes protected repair of existing function and the learning responses left available after repair. Fiber Acquisition Theory (FAT) describes the formation and renewal of trainable structure, distinguishing neutral installation, access to a learning signal, and additional functional directions. The two operations also constrain one another. A repair obligation can force a change in a learning channel, while acquired features can make new protected repairs feasible or reduce the minimum coefficient cost of existing ones. We characterize joint admissibility and response reachability, and give a finite supervised-learning construction in which the two operations together produce a favorable response absent from either operation alone. We implement protected head repair and fixed-capacity LoRA renewal in Qwen2.5-32B with AdamW, under replay and output distillation. Across two training trajectories differing in seed and within-task example order, the coordinated policy achieves higher teacher-forced Exact than either separate policy in all four comparisons. It also has the lowest macro NLL under both learning rules on the second trajectory. The implementations realize structure-conditioned self-regulation within specified information and action constraints: accessible state readouts and arrived evidence guide repairs and structural renewals, and ordinary learning continues from the modified state.
\end{abstract}
\section{Introduction}\label{sec:intro}
A learner's current function does not completely specify how it will learn next. Parameter realization, optimizer memory, trainable representations, and retained experience can affect its response to further training. Observation-relative fibers describe agreement in selected current behavior while controlled future-learning responses distinguish states within that agreement class \citep{wang2026fingerprints}. The central question of this paper is how such structure can be acted upon: how existing function can be repaired, how new learning directions can be formed, and how the two operations reshape each other's possibilities.

We formulate two general theories. \emph{Fiber Regulation Theory} (FRT) characterizes state changes that achieve a functional repair while preserving a declared observation. \emph{Fiber Acquisition Theory} (FAT) characterizes the formation, installation, and use of trainable structure from available evidence, including renewal at fixed capacity. Their shared subject is the evolving learning state; their observations, actions, and immediate objectives may differ.

FRT starts from the finite-frame head repair and intrinsic--extrinsic interaction studied by \citet{wang2026intrinsic}. That construction preserves current logits while repairing selected historical margins; its later effect depends on the continuation. Building on that finite repair family, we characterize the learning responses available after the repair obligation has been met. The new response geometry identifies directions along which later learning can change without undoing the prescribed functional repair.

FAT addresses a different distinction. Installing a new feature with zero output coefficient can leave the current function unchanged while changing the gradient and output response of the next update. We give separate criteria for neutral installation, access to a learning signal, and additional functional directions. A renewal that removes an existing component requires its own finite acceptance conditions. This distinction makes acquisition a property of the subsequent learning dynamics, not merely of parameter count.

When regulation and acquisition act together, their admissible actions and responses can change. Separate acceptance does not generally imply joint acceptance. We characterize a repair obligation that forces a change in a learning channel, derive the protected repair directions available after compensating the current effects of new features, and describe joint response reachability under a norm budget. A finite cross-entropy example establishes strict positive complementarity through an actual supervised gradient update. 

At each decision, the policy uses accessible state readouts and arrived evidence to select admissible repairs or renewals, or to make no extra write. Ordinary learning then proceeds from the resulting state.

\begin{figure}[tbp]
  \centering
\begin{tikzpicture}[
  x=1cm,y=1cm,
  font=\fontsize{9}{11}\selectfont,
  state/.style={draw=figink!75,fill=black!1,line width=.55pt,
    rounded corners=1.5pt,minimum height=.98cm,text width=2.95cm,
    align=center,inner sep=5pt},
  info/.style={draw=figteal,line width=.55pt,fill=figpale,
    rounded corners=1.5pt,minimum height=1.85cm,text width=2.95cm,
    align=center,inner sep=5pt},
  policy/.style={draw=figink!80,fill=white,line width=.65pt,
    rounded corners=1.5pt,text width=5.15cm,minimum height=2.52cm,
    align=center,inner sep=6pt},
  sensing/.style={-{Stealth[length=5pt,width=4pt]},draw=figteal,
    line width=.70pt,dash pattern=on 3pt off 2.5pt},
  intervention/.style={-{Triangle[open,length=6pt,width=5pt]},
    draw=figteal,line width=.80pt},
  learning/.style={-{Triangle[length=6pt,width=5pt]},draw=figink,
    line width=1.15pt},
  feedback/.style={-{Stealth[length=5pt,width=4pt]},draw=figink!65,
    line width=.75pt,densely dotted},
  label/.style={align=center,fill=white,inner sep=2pt}
]
\definecolor{figink}{HTML}{20252A}
\definecolor{figteal}{HTML}{426B73}
\definecolor{figpale}{HTML}{F5F8F8}

\path[use as bounding box] (0,-5.12) rectangle (16,1.05);
\node[text=figink] at (8,.98) {\textbf{One evolving learner}};

\node[state] (state) at (1.72,0) {Complete state\\[2pt]$s_t$};
\node[state] (acted) at (8,0) {After intervention\\[2pt]$\widetilde s_t$};
\node[state] (learned) at (14.28,0) {After learning\\[2pt]$s_{t+1}$};

\draw[intervention] (state.east) -- node[label,above=3pt]
  {$\overline K_{\xi_t}$} (acted.west);
\draw[learning] (acted.east) -- node[label,above=3pt]
  {Ordinary learning} (learned.west);

\node[info] (info) at (1.72,-2.75)
  {\textbf{Legal information}\\[3pt]
   $i_t$,\quad $\sigma(i_t)=\mathcal I_t$\\[4pt]
   Partial state readouts\\
   Arrived evidence};
\node[policy] (policy) at (8,-2.75)
  {\textbf{Policy and admission}\\[4pt]
   \mbox{FRT (repair) $\longleftrightarrow$ FAT (form / renew)}\\[2pt]
   Possible cross-effects\\[5pt]
   $\xi_t=(f_t,a_t)\in\mathscr K(s_t)$\\[4pt]
   Identity choices allowed};
\node[info] (nextinfo) at (14.28,-2.75)
  {\textbf{Next information}\\[3pt]
   $i_{t+1}$\\[4pt]
   Partial legal readouts\\
   Arrived evidence};

\draw[sensing] (state.south) -- node[label,left=4pt] {Sensing} (info.north);
\draw[sensing] (learned.south) -- node[label,right=4pt] {Sensing} (nextinfo.north);
\draw[sensing] (info.east) -- (policy.west);

\coordinate (commit) at ($(state.east)!.50!(acted.west)$);
\draw[intervention] (policy.north) -- (8,-.88) -| (commit);

\draw[feedback] (nextinfo.south) -- (14.28,-4.42)
  -- node[label] {Next decision: $t\leftarrow t+1$} (1.72,-4.42)
  -- (info.south);

\begin{scope}[shift={(1.10,-5.00)}]
  \draw[sensing] (0,0)--(.58,0);
  \node[anchor=west] at (.70,0) {Information};
  \draw[intervention] (3.50,0)--(4.08,0);
  \node[anchor=west] at (4.20,0) {Intervention};
  \draw[learning] (7.25,0)--(7.83,0);
  \node[anchor=west] at (7.95,0) {Learning};
  \draw[feedback] (10.72,0)--(11.30,0);
  \node[anchor=west] at (11.42,0) {Feedback};
\end{scope}
\end{tikzpicture}
  \caption{Regulation and acquisition within one evolving learner.
  The complete state $s_t$ is distinct from the available information $i_t$.
  Regulation repairs existing function; acquisition forms or renews trainable
  structure. Using $\mathcal I_t$, a policy selects an admissible pair
  $\xi_t=(f_t,a_t)\in\mathscr K(s_t)$ through available certificates or a valid robust rule.
  Identity components allow either intervention, both, or no extra write.
  The state-transition marginal $\bar K_{\xi_t}$ uses the declared execution order.
  Ordinary learning consumes the resulting state and supplies readouts to later decisions.
  The FRT--FAT arrow denotes possible cross-effects in admissibility, response and repair cost.}
  \label{fig:state-regulation-acquisition}
\end{figure}
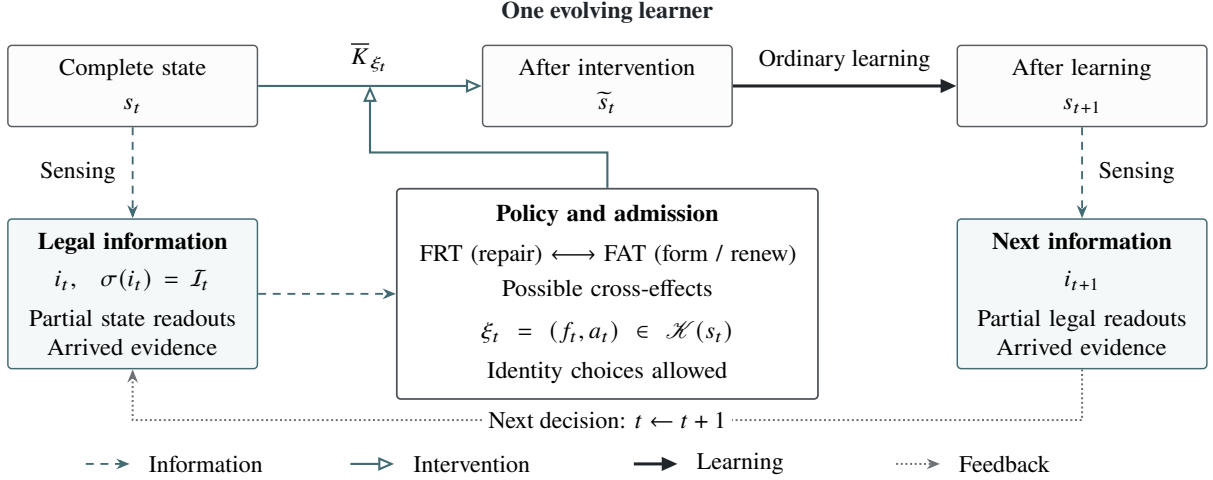

Protected updates and preservation--editing constraints have precedents in orthogonal gradient descent, constrained model editing, and null-space editing \citep{farajtabar2020ogd,gupta2024unified,fang2025alphaedit}. Function-preserving expansion, selective plasticity renewal, low-rank budget allocation, and gradient-aligned initialization also have established formulations \citep{chen2016net2net,dohare2024plasticity,zhang2023adalora,wang2024loraga}. Recent low-rank analyses address composite-update interference and parameterization-dependent optimization \citep{chen2026janus,castin2026balanced}. The contribution here is the mathematical relation between the two dynamical operations: residual learning freedom after repair, compensation-adjusted repair directions supplied by acquisition, and the constraints governing their joint execution and response.

After the preliminaries in Section~\ref{sec:prelim}, Sections~\ref{sec:frt}--\ref{sec:coupling} develop the three theoretical components. Section~\ref{sec:llm} derives their transformer/LLM specializations. Section~\ref{sec:evidence} examines Qwen2.5-32B under replay and output distillation across two training trajectories. The coordinated policy achieves higher Exact than either isolated mechanism in all four comparisons. These comparisons show that the specified repairs and renewals can be executed in a trained language model and can affect its measured outcomes after continued learning.

\section{Mathematical preliminaries}\label{sec:prelim}
We introduce only the objects shared by regulation and acquisition. Let $s\in\mathcal S$ be a complete learning state. It contains the parameters and all retained quantities needed for subsequent execution: optimizer state, structural configuration, relevant memory or teacher state, stream position, random state, and any policy or resource state that affects future operations. Different configurations may have different parameter dimensions; formally, $\mathcal S$ may be a disjoint union $\coprod_\rho\mathcal S_\rho$. An observation is not assumed to retain this whole state.

\begin{definition}[Observation-relative fiber]\label{def:fiber}
For a declared observation map $B:\mathcal S\to\mathcal Y$, the fiber through $s$ is
\begin{equation}
 \mathcal F_B(s)=\{\tilde s:B(\tilde s)=B(s)\}.\label{eq:fiber}
\end{equation}
For a stochastic observation, equality refers to its declared conditional law, not to two coincident sample values. On a fixed differentiable chart, $\ker DB_s$ is the infinitesimally invisible subspace.
\end{definition}
The set in \eqref{eq:fiber} needs no manifold assumption. An infinitesimally invisible direction in $\ker DB_s$ need not yield an exactly preserving finite displacement. Conversely, a function-neutral structural extension can be compared through $B$ even when it changes configuration. The term \emph{hidden learning structure} refers here to distinctions erased by the selected observation but relevant to permitted future learning responses; it does not denote a separately autonomous subsystem \citep{wang2026fingerprints,wang2026intrinsic}.

For probabilistic statements, the state and outcome spaces are standard Borel spaces and all kernels and policies are measurable. At an intervention time, $\mathcal I_t$ denotes the sigma-field generated by information legally available before the decision, including any authorized randomization already drawn. \emph{Intrinsic sensing} means a specified readout of this information and state interface, not free access to the complete state or to future outcomes. A lawful operation $a$ has a declared domain and a normalized joint outcome--successor kernel $K_a(s;dy,ds')$ on that domain. Its state-transition marginal is $\bar K_a(s,ds')=K_a(s;\mathcal Y_a,ds')$. Any returned outcome or cost needed for later decisions is retained in $s'$. Deterministic writes are partial maps. An unavailable operation has no probability row; an allowed rejection, deferment, or termination is an actual outcome. A no-intervention branch skips the additional write, not the ordinary learning that follows. Costs and retained diagnostic effects belong to the comparison when relevant.

A continuation $\pi$ fixes the subsequent learning policy, horizon convention, output semantics, and exogenous randomization. Write $P_H^\pi(s,\cdot)$ for the resulting endpoint or path law. The response fingerprint is the family of such laws over a declared probe class, not an automatically available online vector. For an integrable common readout $J$, define
\begin{equation}
 V_H^\pi(s,a;J)
 =\int J\,d\bigl[(\bar K_aP_H^\pi)(s)-(\bar K_0P_H^\pi)(s)\bigr].\label{eq:value}
\end{equation}
Here $\bar K_0$ is the state-transition kernel of the stated reference branch at the same phase cut. The notation includes any prescribed installation and consumption before the continuation. Comparisons across structural dimensions use a common readout or output law, not subtraction of differently typed parameter vectors.

For matched deterministic branches with the identity as reference, \eqref{eq:value} reduces to
$J(\Phi_H^{\pi,\omega}(a(s)))-J(\Phi_H^{\pi,\omega}(s))$.
The same $\omega$ means the same declared external realization, not merely equal seed integers with incompatible random-number consumption. A shared feedback policy may issue different later actions after its states diverge.

Throughout the paper, \emph{local admissibility}, \emph{continuation-conditioned value}, and \emph{complete-policy performance} are different claims. A current certificate establishes the first. Equation~\eqref{eq:value} defines the second. Comparing two live policies from a common initial law addresses the third. An Exact score is a discrete readout; smooth calculations below apply to stated losses or response maps, never to an unqualified derivative of Exact.

\paragraph{Linear-algebra conventions.}
All main finite-dimensional statements use Euclidean inner products, Frobenius matrix norms, and the Moore--Penrose inverse $M^\dagger$. The operator norm is $\norm{M}_{\rm op}$. Projectors are orthogonal. Any restriction to an implemented actuator must be included in the action coordinates before applying a theorem. In particular, an ideal dense action space is not automatically an implementation's available action set.

\section{Fiber Regulation Theory}\label{sec:frt}
\subsection{Regulation as a constrained change of learning state}
A regulation problem specifies what is protected, what needs repair, and which actions can be executed. Its target is a functional obligation on an existing learner; it need not preserve every parameter or every future response.

\begin{definition}[Regulation specification]\label{def:regulation}
At a fixed decision state $s$, let $B_s$ be the declared protection observation, $R_s:\mathcal S\to\mathcal Y_R$ the repair-target readout, $\mathcal O_s\subseteq\mathcal Y_R$ its acceptable values, $\mathcal A_F(s)$ an executable action interface, and $c_F(s,f)$ a cost or size constraint. A successful regulation action satisfies
\begin{equation}
 f\in\mathcal A_F(s),\qquad B_s(f(s))=B_s(s),\qquad R_s(f(s))\in\mathcal O_s,\qquad c_F(s,f)\le b_F.\label{eq:frt-def}
\end{equation}
The proposal and admission decision are $\mathcal I_t$-measurable. A registered no-write branch remains available when no proposal is accepted, without being counted as a successful repair.
\end{definition}
Finite-precision implementations replace exact constraints by declared tolerances checked on the stored action. The subscript $s$ fixes the consultation's frames and meanings while its alternatives are compared. It does not permit changing the observation after seeing which action succeeds. FRT is the study of this regulated evolution and its learning consequences, not a requirement to intervene at every update.

\subsection{Exact finite repair geometry}
An important class is affine in the action even when the underlying predictor is nonlinear. Let $X\in\R^{d\times q}$ be a permitted action coordinate, $C\in\R^{m\times d}$ the protected response matrix, $H\in\R^{p\times d}$ the obligation matrix, and $D\in\R^{p\times q}$ the desired change. The finite repair equations are
\begin{equation}
 CX=0,\qquad HX=D.\label{eq:repair}
\end{equation}
This notation includes the independently writable head slice of \citet{wang2026intrinsic}, but does not require a language model.

\begin{theorem}[Finite protected repair and remaining freedom]\label{thm:repair}
Set $P_C=I-C^\dagger C$ and $N=HP_C$. Equations~\eqref{eq:repair} are feasible if and only if
\begin{equation}
 NN^\dagger D=D.\label{eq:repair-feas}
\end{equation}
When feasible, the unique minimum-Frobenius-norm solution is $X_*=N^\dagger D$. Every solution is
\begin{equation}
 X=X_*+Z,\qquad CZ=HZ=0,\qquad
 \norm{X}_F^2=\norm{X_*}_F^2+\norm{Z}_F^2.\label{eq:repair-all}
\end{equation}
Consequently a norm budget $\tau$ permits a repair exactly when \eqref{eq:repair-feas} holds and $\norm{X_*}_F\le\tau$.
\end{theorem}
\begin{proof}
The first constraint means $X=P_CX$. Hence $HX=NX$, whose solvability is \eqref{eq:repair-feas}. The minimum-norm solution of $NX=D$ is $N^\dagger D$. Since $\Ran N^\top\subseteq\Ran P_C$, this solution also satisfies $CX_*=0$. The difference of any two solutions lies in the common kernel of $C$ and $H$. Every column of $X_*$ lies in $\Ran N^\top$, orthogonal to the columns of such a difference. This proves the norm identity and uniqueness.
\end{proof}
The complete affine solution family and norm budget are inherited from \citet[Corollary~3.2]{wang2026intrinsic}; they are recalled to define the residual space used in Theorem~\ref{thm:authority}. It does not assert that the physical actuator can execute every $Z$ in \eqref{eq:repair-all}. When it can, the same protected repair can have multiple learning realizations. When it cannot, its narrower action interface must be used.

\subsection{Learning responses remaining after the same repair}
The repair equations specify present functional effects. They do not generally fix the learning response. To make the distinction quantitative, work in a finite-dimensional Hilbert action space $\mathcal U$. Let $\mathcal Z\subseteq\mathcal U$ be the linear space of variations that preserve both the protection and the completed repair, and let $\Pi_{\mathcal Z}$ be its projector. Let $L:\mathcal U\to\mathcal V$ describe a declared linear learning-response channel at the fixed parent. It can be an exact finite channel in an affine slice or a first-order channel on a qualified differentiable branch; these are different interpretations.

\begin{theorem}[Repair-compatible response geometry]\label{thm:authority}
Suppose the residual norm budget in \eqref{eq:repair-all} is $\rho\ge0$. The additional reachable responses are
\begin{equation}
 \mathcal E_{\mathcal Z}(\rho)
 =\{LZ:Z\in\mathcal Z,\ \norm{Z}\le\rho\}
 =\{L\Pi_{\mathcal Z}u:\norm{u}\le\rho\}.\label{eq:authority-set}
\end{equation}
Their squared directional support is determined by
\begin{equation}
 G_{\mathcal Z}=L\Pi_{\mathcal Z}L^*,\qquad
 \sup_{v\in\mathcal E_{\mathcal Z}(\rho)}\ip{w}{v}
 =\rho\sqrt{\ip{w}{G_{\mathcal Z}w}}.\label{eq:authority-gram}
\end{equation}
For $\rho>0$, the channel is identically zero exactly when $L\Pi_{\mathcal Z}=0$. If added protection reduces $\mathcal Z$ to $\mathcal Z'\subseteq\mathcal Z$ while $L$ is fixed, then
\begin{equation}
 0\preceq G_{\mathcal Z'}\preceq G_{\mathcal Z}.\label{eq:protection-monotone}
\end{equation}
\end{theorem}
\begin{proof}
Projection does not increase norm and every $Z\in\mathcal Z$ is its own projection, giving the set identity. The support maximization is
$\sup_{\norm{u}\le\rho}\ip{\Pi_{\mathcal Z}L^*w}{u}
=\rho\norm{\Pi_{\mathcal Z}L^*w}$.
Squaring the norm gives \eqref{eq:authority-gram}. The zero-channel statement follows by scaling any nonzero image direction into the ball. Nested orthogonal projections satisfy
$\Pi_{\mathcal Z'}\preceq\Pi_{\mathcal Z}$; congruence by $L$ proves \eqref{eq:protection-monotone}.
\end{proof}
A large observation kernel can therefore have zero image in the chosen learning-response channel. Conversely, the same repair may leave a nonzero family of such responses. The monotonicity statement is a \emph{fixed-parent} comparison: after a real repair changes the parent and hence $L$, it does not order the new and old Gramians automatically. This prevents geometric freedom from being mistaken for a scalar measure of overall learning quality.

\subsection{Output protection need not protect the backward learning channel}
Let a differentiable representation $h_\theta(x)\in\R^d$ be followed by an independently writable head $W\in\R^{q\times d}$, with logits $z=Wh_\theta(x)$. At a fixed example, write $J_\theta=D_\theta h_\theta(x)$ and $r=\nabla_z\ell(z,y)$. A head action is $W\mapsto W+X^\top$. Its generation is held fixed when computing the gradient consumed by subsequent learning; this is not differentiation through a meta-generator.

\begin{proposition}[Protected-output backward response]\label{prop:backward}
If $X^\top h_\theta(x)=0$, then the logits, loss, and $r$ at that example are unchanged, but
\begin{equation}
 \Delta_F\nabla_\theta\ell=J_\theta^\top Xr.\label{eq:backward}
\end{equation}
For a fixed collection of examples and fixed sample weights, the corresponding expression is the weighted sum of \eqref{eq:backward}, provided $X^\top h_\theta(x_i)=0$ for \emph{every} contributing example $i$. The identity is finite and exact in $X$.
\end{proposition}
\begin{proof}
Before the write, the gradient is $J_\theta^\top W^\top r$; after it, it is $J_\theta^\top(W^\top+X)r$. The equality of logits makes the same $r$ valid in both expressions. Subtraction gives the result.
\end{proof}
This is the protected-backpropagation identity in Appendix~A.4 of \citet{wang2026intrinsic}: present functional neutrality and future learning neutrality are different constraints. It supplies a concrete $L$ for Theorem~\ref{thm:authority}. Its exactness does not bypass numerical execution checks, nor does it imply that an arbitrary optimizer converts the changed gradient into a beneficial parameter step.

\subsection{Regulation and external learning}
For one fixed regulation proposal $f$, two continuations $\pi_0,\pi_1$, and common readout $J$, write $Y_{ij}=\E[J_{ij}]$ for the expected readout with $f$ absent/present ($i$) and continuation $\pi_j$; in deterministic matched branches, the expectation reduces to the recorded readout. Then
\begin{equation}
 I_{F,\pi}=Y_{11}-Y_{01}-Y_{10}+Y_{00}
 =V_H^{\pi_1}(s,f;J)-V_H^{\pi_0}(s,f;J).\label{eq:frt-ie}
\end{equation}
This is the intrinsic--extrinsic coupling identified by the matched design of \citet{wang2026intrinsic}. It asks whether continuation changes the marginal value of the \emph{same} internal action. The equivalent reading, that the action changes the continuation's incremental value, is the same contrast, not a reverse-time intervention. A complete regulation policy can subsequently use a state signal to choose later experience allocation, giving an additional, genuinely temporal feedback loop.

Thus FRT contains both a geometry of admissible repair and a dynamics of how those repairs are consumed by continued learning. The former determines what the intervention can guarantee immediately; the latter determines its continuation-specific role.

\section{Fiber Acquisition Theory}\label{sec:fat}
\subsection{Formation, installation, and consumption}
Regulation starts from a functional obligation. Acquisition starts from available experience and the possibility of changing the learner's trainable realization. The two problems share a state and a response language, but not necessarily the same observation, admissible action family, or target.

\begin{definition}[Acquisition operation]\label{def:acquisition}
Let $i_t$ be the permitted information record generating $\mathcal I_t$, and let $u$ be a fresh authorized random variable, independent of unobserved future exogenous information conditional on $\mathcal I_t$. An acquisition specification first declares a family of trainable components and an installation interface that creates, replaces, or changes their trainable realization. The index $v$ refers to that interface. Identity and HOLD are control branches; an ordinary consumer update is not classified as acquisition merely by relabeling it as installation. An acquisition operation consists of a formation rule $v=\Gamma(i_t,u)$ measurable with respect to $\mathcal I_t\vee\sigma(u)$, an admission predicate using that same information, a lawful installation $A_v$, and a specified learning rule (the \emph{consumer}) $U_c$. State-dependent formation reads only the components included in $i_t$; $c$ specifies the update mechanism. The resulting state change is assessed through
\begin{equation}
 s\longmapsto A_v(s)\longmapsto U_c(A_v(s)),\label{eq:fat-operation}
\end{equation}
including its persistent structure, optimizer state, returned outcomes, and costs. A rejected installation follows its registered no-installation branch. Continued ordinary learning is not disabled.
\end{definition}
The generator may use a current structural readout, already encountered evidence, and finite paid diagnostics. The definition does not require a forecast of an unobserved task's value. Acquisition is established operationally by installation and use; favorable task value is a further property. In particular, a new tensor that no future permitted computation can consume is not evidence of acquired learning capability.

There are two important interfaces. An \emph{extension} preserves the old realization and adds new degrees of freedom. A \emph{renewal} changes the realization within a fixed capacity, possibly removing an old component. An extension may be function-neutral at installation; a renewal may require a finite damage budget. Neither interface is defined by its parameter count alone.

\subsection{Function-neutral birth and accessible learning}
Consider a family of predictors with an installed feature map $\phi_v(x)\in\R^k$ and new coefficients $A\in\R^{q\times k}$:
\begin{equation}
 f_{v,A}(x)=f_0(x)+A\phi_v(x),\qquad A\big|_{\rm install}=0.\label{eq:neutralbirth}
\end{equation}
For a fixed batch of $n$ examples, put their feature rows in $\Phi_v\in\R^{n\times k}$, their output-gradient rows in $D\in\R^{n\times q}$, and absorb the declared sample weights into $D$. The features are held fixed for the following single-step statement. The loss has the output-composed form $L(A)=\ell(Z_0+\Phi_vA^\top)$, with $D=\nabla_Z\ell(Z_0)$; it has no additional direct parameter penalty.

\begin{theorem}[Neutral installation, learning-active structure]\label{thm:birth}
All choices of $v$ in \eqref{eq:neutralbirth} compute the same current function at installation. Their coefficient gradients are nevertheless
\begin{equation}
 \nabla_A L\big|_{A=0}=D^\top\Phi_v.\label{eq:birthgrad}
\end{equation}
If the consumer performs one SGD step on $A$ alone with rate $\eta>0$, keeping $f_0$ and $\Phi_v$ fixed, the exact change of the batch output matrix is
\begin{equation}
 \Delta Z_v=-\eta\Phi_v\Phi_v^\top D.\label{eq:birthkernel}
\end{equation}
For two installed structures $v,w$, the resulting \emph{one-step batch-output matrices differ} exactly when
\begin{equation}
 (\Phi_v\Phi_v^\top-\Phi_w\Phi_w^\top)D\ne0.\label{eq:birthdistinct}
\end{equation}
\end{theorem}
\begin{proof}
The added output is zero at $A=0$ for every input. Differentiating $\Phi_v A^\top$ gives $D^\top\Phi_v$. The stipulated update is $A^+=-\eta D^\top\Phi_v$, so the new batch output is $\Phi_v(A^+)^\top=-\eta\Phi_v\Phi_v^\top D$. Subtracting two such outputs proves the last assertion.
\end{proof}
This theorem identifies a concrete hidden learning distinction: an unchanged function can have a changed accessible learning operator. The finite-feature response $\Phi_v\Phi_v^\top$ is a kernel gradient operator of the familiar finite-feature type \citep{jacot2018ntk}. Function-preserving expansion itself is established prior work \citep{chen2016net2net,hu2021lora}. The relevant acquisition question is what response the new structure makes possible under its actual consumer. Equation~\eqref{eq:birthkernel} is not an exact formula for simultaneously training all parameters or for AdamW; those consumers have different response maps. Equation~\eqref{eq:birthgrad} remains the corresponding current coefficient-gradient identity.

\subsection{Protected access to a current learning signal}
A useful specialization forms a scalar feature $a^\top x$ from a currently available input frame $X\in\R^{n\times d}$, with a newborn output vector $b\in\R^q$ initialized to zero. Let $P$ project onto a declared protected or retained input span, and let $D\in\R^{n\times q}$ be the current output-loss gradient. The new module contributes $b(a^\top x)$; the loss depends on this module through its output, as in Theorem~\ref{thm:birth}.

\begin{proposition}[Protected current-signal direction]\label{prop:birthdirection}
For unit $a\in\ker P$, the newborn gradient is $\nabla_bL=D^\top Xa$. Let
\begin{equation}
 G=D^\top X(I-P).\label{eq:projected-signal}
\end{equation}
If $G\ne0$, a leading right singular vector of $G$ with nonzero singular value lies in $\ker P$ and attains
\begin{equation}
 \max_{\norm{a}=1,\,Pa=0}\norm{D^\top Xa}=\norm{G}_{\rm op}.\label{eq:singularmax}
\end{equation}
If $G=0$, no direction in this protected complement has a nonzero newborn gradient for the stated frame and loss.
\end{proposition}
\begin{proof}
The chain rule gives the gradient. For $Pa=0$, $D^\top Xa=Ga$. The top singular value is the largest norm over unit inputs. Since $G^\top G=(I-P)G^\top G(I-P)$, an eigenvector with positive eigenvalue lies in $\ker P$ and attains the restricted maximum. The zero case is immediate.
\end{proof}
The singular-vector construction is a standard variational fact, related to gradient-aligned low-rank initialization \citep{wang2024loraga}. The theorem's content for FAT is the explicit intersection of \emph{current protection} and \emph{current learning access}. It neither forecasts the optimal future direction nor makes the instantaneous gradient norm a utility function.

The same formula clarifies an internal interaction between structural constraints and arriving evidence. In a comparison where the captured signal $H_e=D_e^\top X_e$ and projector $P_i$ are independently specified on a common domain, let $G(P_i,e)=H_e(I-P_i)$. Its mixed difference is exactly
\begin{equation}
 \begin{split}
 &G(P_1,e_1)-G(P_0,e_1)-G(P_1,e_0)+G(P_0,e_0)\\
 &\hspace{20mm}=-(H_{e_1}-H_{e_0})(P_1-P_0).\label{eq:fat-internal}
 \end{split}
\end{equation}
This gives a conditional operator-level nonadditivity criterion. In a live learner, evidence, features, projectors, and optimizer memory can all depend on the parent. Equation~\eqref{eq:fat-internal} therefore cannot substitute for a properly crossed empirical comparison, and nonzero interaction does not imply positive value.

\subsection{Functional novelty is distinct from parameter novelty}
At a fixed differentiable state and a fixed common output frame, let $J_o$ map the old trainable coordinates to output variations and $J_n$ map newly installed coordinates. Both are finite matrices in this statement. The relative-image construction uses the same residual-image principle as the observation-relative response geometry of \citet{wang2026fingerprints}, here applied to old and newly writable coordinates.

\begin{proposition}[Additional tangent response dimension]\label{prop:novelty}
The number of linearly independent output directions added by the new coordinates, relative to the old tangent image, is
\begin{equation}
 \nu=\rank[J_o\ J_n]-\rank J_o
 =\rank\bigl[(I-J_oJ_o^\dagger)J_n\bigr].\label{eq:novelty}
\end{equation}
It is zero exactly when $\Ran J_n\subseteq\Ran J_o$.
\end{proposition}
\begin{proof}
Resolve $\Ran J_n$ into its component in $\Ran J_o$ and its orthogonal complement. The latter is the image of $(I-J_oJ_o^\dagger)J_n$, and its dimension is exactly the increment in the dimension of the summed image.
\end{proof}
Three distinctions follow. A nonzero new parameter may have zero current functional derivative. A nonzero derivative may lie in the old tangent image. Finally, even zero additional tangent dimension need not imply identical optimization dynamics: coordinate realization and optimizer state can change how an existing function space is traversed \citep{castin2026balanced}. Thus \eqref{eq:novelty} is a finite-frame, first-order novelty criterion, not the dimension of all future capabilities or a requirement for every useful renewal.

\subsection{Finite protection beyond zero output}
Neutral installation by itself controls only the installation instant. One can additionally bound the effect of a prescribed nonzero use of the new structure on a current representation frame. Let $H\in\R^{n\times d}$, $n\ge1$, be that frame, $\Phi\in\R^{n\times k}$ the installed features, and $U\in\R^{d\times k}$ a prospective output coefficient with $\norm{U}_F\le\rho$. Put $E=\Phi U^\top$ and $H^+=H+E$.

\begin{proposition}[Finite geometric budget]\label{prop:gram}
The row-Gram change is exactly
\begin{equation}
 H^+(H^+)^\top-HH^\top
 =HU\Phi^\top+\Phi U^\top H^\top+\Phi U^\top U\Phi^\top.\label{eq:gramidentity}
\end{equation}
Writing $e=\rho\norm{\Phi}_{\rm op}$ gives
\begin{equation}
 \norm{H^+(H^+)^\top-HH^\top}_F
 \le 2\norm{H}_{\rm op}e+e^2.\label{eq:grambound}
\end{equation}
For $C_H=H^\top H/n$ and $S_\tau(C)=C(C+\tau I)^{-1}$, $\tau>0$,
\begin{equation}
 \norm{S_\tau(C_{H^+})-S_\tau(C_H)}_{\rm op}
 \le \frac{2\norm{H}_{\rm op}e+e^2}{n\tau}.\label{eq:softspectral}
\end{equation}
\end{proposition}
\begin{proof}
Expansion gives \eqref{eq:gramidentity}. Since $\norm{E}_F\le e$, each cross term has Frobenius norm at most $\norm{H}_{\rm op}e$ and $\norm{EE^\top}_F\le e^2$. The analogous expansion of $(H+E)^\top(H+E)$ bounds its operator norm by the same numerator. Finally $S_\tau(C)=I-\tau(C+\tau I)^{-1}$. The resolvent identity and $\norm{(C+\tau I)^{-1}}_{\rm op}\le1/\tau$ for positive semidefinite $C$ yield the stated bound.
\end{proof}
These bounds concern the declared frame, insertion point, and coefficient budget. If existing residual components change the true base frame from $H$ to $H+E_o$, the bound must be recomputed or transported with a bound on $E_o$; certifying $H$ does not certify the new base automatically. This finite qualification is compatible with later free learning, but it is not a perpetual protection theorem. The final fixed-capacity LoRA case study below uses a different, directly evaluated admission interface; it does not claim to implement every spectral certificate.

\subsection{The actual learning consumer matters}
A candidate may alter a gradient, an installed representation, optimizer memory, or several of these together. A gradient summary describes the whole learning effect only under an additional sufficiency condition.

\begin{theorem}[Local consumer sufficiency]\label{thm:consumer}
On a fixed differentiable action chart $a\in\R^k$ at $a=0$, let $L:\R^k\to\R^p$ be the derivative of a selected gradient readout and $F_c:\R^k\to\R^m$ the derivative of a common coordinate representation of the complete successor under consumer $c$. There exists a linear $R_c$ such that $F_c=R_cL$ if and only if
\begin{equation}
 \ker L\subseteq\ker F_c.\label{eq:consumerfactor}
\end{equation}
For a downstream differentiable readout with derivative $P_H$, the weaker factorization $P_HF_c=R_{c,H}L$ holds exactly when $\ker L\subseteq\ker(P_HF_c)$.
\end{theorem}
\begin{proof}
A factorization makes every zero direction of $L$ a zero direction of the factored map. Conversely, define $R_c(La)=F_ca$ on $\Ran L$. The kernel inclusion makes this independent of the representative $a$; extend linearly to the rest of the finite-dimensional space. The downstream statement applies the same argument to $P_HF_c$.
\end{proof}
For example, $L(a_1,a_2)=a_1$ and $F_c(a_1,a_2)=(a_1,a_2)$ violate \eqref{eq:consumerfactor}. The same gradient summary can conceal a persistent structural change. Even a valid derivative factorization need not be a finite one: $L(a)=0$ and a successor coordinate $a^2$ have zero derivatives at zero but different finite successors.

The corresponding exact finite statement is simple and useful. If two actions induce the same complete successor law, including retained state and costs, then every subsequent common lawful policy induces the same output law. Equality of one gradient or one endpoint score is weaker. Appendix~\ref{app:consumer} gives a concrete optimizer-coordinate example. FAT therefore treats the consumer as part of the dynamical object rather than assuming that the same raw direction has the same meaning under SGD, AdamW, replay, and distillation.

\subsection{Renewal under bounded capacity}
For a factorized module $M=\sum_{j=1}^r b_ja_j^\top$, replacing one physical slot by $(\tilde a_j,0)$ gives
\begin{equation}
 M^+=M-b_ja_j^\top.\label{eq:turnover}
\end{equation}
The newborn component is output-zero; removal of the old component need not be. Full replacement must therefore satisfy its own finite admissibility checks. Two different accepted newborn directions can nevertheless give the same $M^+$ while producing different future updates through \eqref{eq:birthgrad}. The updated optimizer slices are part of the renewed state.

Continual backpropagation also renews selected units by resetting incoming weights and initially zeroing outgoing weights \citep{dohare2024plasticity}. Low-rank factor resetting and optimizer-state adjustments appear in ReLoRA, which merges accumulated low-rank updates into a base weight before reinitialization \citep{lialin2024relora}. The present interface instead removes one physical slot and admits the new direction only after finite protection checks.

This distinguishes the mathematics of \emph{forming a learning-active realization} from a blanket requirement for exact current output preservation. FRT and FAT share the fiber viewpoint, but FAT need not inherit every FRT protection equation. Their compatibility is a separate joint problem.

\section{Coupled regulation--acquisition dynamics}\label{sec:coupling}
\subsection{Cross-response, not merely co-location}
FRT and FAT are two action interfaces on one evolving learner. They are not assumed to partition that learner into disjoint physical subsystems. Let $T_F^f$ and $T_A^a$ denote two lawful finite writes at a common parent, and let $O_F,O_A$ be declared regulation and acquisition readouts. For numerical components, define
\begin{equation}
 \Delta_j O_i(s;a_j)=O_i(T_j^{a_j}s)-O_i(s),\qquad i,j\in\{F,A\}.\label{eq:crossresponse}
\end{equation}
The collection
\begin{equation}
 \mathcal R(s)=
 \begin{pmatrix}
 \Delta_F O_F&\Delta_A O_F\\
 \Delta_F O_A&\Delta_A O_A
 \end{pmatrix}\label{eq:fourchannels}
\end{equation}
is a typed response table, not necessarily a numerical $2\times2$ matrix on which multiplication is meaningful. A candidate set, a rank label, or an accept/reject outcome is compared as a set or label, not by an invented numerical difference.

The off-diagonal entries express directional cross-effects. A nonzero acquisition-gradient response after a regulation action is one such effect; a changed repair residual or feasible set after acquisition is another. Merely sharing parameters does not establish either. Nor do internal FRT--continuation and FAT--evidence interactions logically force nonzero cross-effects between FRT and FAT. Coordination is the policy that uses these effects; coupling is the underlying state-dependent response relation.

\subsection{Joint admissibility and sequential execution}
Fix candidate labels, actual decoders, state version, and execution order. Let $\mathscr K(s)$ be the relation of candidate pairs whose \emph{joint execution} satisfies the stated current contract.

\begin{proposition}[Nonrectangular joint admissibility]\label{prop:rectangular}
Two independent qualification sets represent all and only jointly admissible pairs if and only if
\begin{equation}
 \mathscr K(s)=\operatorname{proj}_F\mathscr K(s)
             \times\operatorname{proj}_A\mathscr K(s).\label{eq:rectangular}
\end{equation}
For partial deterministic commit maps $F,A$ with domains $D_F,D_A$,
\begin{equation}
 \dom(F\circ A)=D_A\cap A^{-1}(D_F),\qquad
 \dom(A\circ F)=D_F\cap F^{-1}(D_A).\label{eq:partialdomain}
\end{equation}
Thus separate qualification at the parent does not establish either composition.
\end{proposition}
\begin{proof}
Any nonempty product relation is the product of its projections; conversely that product is represented by its two projections. The empty case is also represented by empty qualification sets. Equation~\eqref{eq:partialdomain} is precisely the condition that the first map exists and its result belongs to the second map's domain.
\end{proof}
Nonrectangularity can express conflict or compensation. For instance, the pairs $(0,0)$ and $(1,1)$ alone form a nonrectangular set; knowing that each label appears in some valid pair does not authorize $(0,1)$. A repair can also make an acquisition admissible that was inadmissible by itself. Consequently a joint solver or a finite ordered recertification must use the relation itself, not assume that two isolated certificates compose.

There is an equally important information condition. For a finite candidate universe, an observation $O(s)$ determines the exact relation precisely when
\begin{equation}
 O(s)=O(\tilde s)\ \Longrightarrow\ \mathscr K(s)=\mathscr K(\tilde s).\label{eq:joint-info}
\end{equation}
A robust policy using only $O$ must instead select from the intersection over all compatible states. The proof and the distinction between lawful HOLD and nontrivial progress are given in Appendix~\ref{app:information}. No full-state information is granted by naming a joint relation.

\subsection{Exact finite cross-effects on function and certificates}
In an affine output slice, let $C$ contain feature rows, $W$ the output head, $X$ a head increment in transposed coordinates, and $E$ an acquisition-induced change of those feature rows. Fixed head and feature interventions have the combined output difference
\begin{equation}
 (C+E)(W^\top+X)-CW^\top=CX+EW^\top+EX.\label{eq:mixedlogits}
\end{equation}
Even if $CX=0$ and $EW^\top=0$ separately, the mixed term $EX$ can remain. For example, $C=(1,0)$, $W^\top=(1,0)^\top$, $X=(0,1)^\top$, and $E=(0,1)$ give two individually zero output changes and a joint change of one. This finite algebraic interaction does not require noncommuting state edits; the edits of independent coordinates in this example commute.

The same calculation transports a regulation certificate. Suppose its residuals are
$r_C=CX$ and $r_H=HX-D$. If an acquisition changes the frames and obligation to
$C'=C+E_C$, $H'=H+E_H$, $D'=D+\Delta D$, while the old write $X$ is held fixed, then
\begin{equation}
 r_C'=r_C+E_CX,\qquad r_H'=r_H+E_HX-\Delta D.\label{eq:residualtransport}
\end{equation}
Equation~\eqref{eq:residualtransport} is exact, not a first-order approximation. It explains why a certificate is state-version-specific. It does not predict the residual of a newly recomputed canonical write, and it does not identify all future forgetting with these finite-frame residuals.

\subsection{When repair must alter a learning channel}
Cross-effects need not be accidental. A repair obligation can be incompatible with preserving an acquisition channel.

\begin{theorem}[Obligation--learning-channel conflict]\label{thm:forced}
At a fixed representation state, let $J\in\R^{d\times k}$ be the Jacobian of a chosen acquisition coordinate. Suppose the historical frame satisfies
\begin{equation}
 H=LC+BJ^\top\label{eq:overlap}
\end{equation}
for some matrices $L,B$. Every write satisfying $CX=0$ and $HX=D$ then satisfies
\begin{equation}
 D=BJ^\top X.\label{eq:forced}
\end{equation}
If the repair is feasible and $D\ne0$, then $\norm B_{\rm op}>0$, no such write has $J^\top X=0$, and
$\norm{J^\top X}_F\ge\norm{D}_F/\norm{B}_{\rm op}$.
For the all-label statement, let $J=D_\theta h_\theta(x)$ at an example $x$ represented in the protected frame $C$, and use cross-entropy $\ell_y(z)=-\log(\softmax z)_y$ for a $q$-class head. With a centered write $X\mathbf1=0$, preserving this example's coordinate gradient for every possible label is equivalent to $J^\top X=0$. Thus nonzero $D$ under \eqref{eq:overlap} forbids that stronger gradient-preservation demand.
\end{theorem}
\begin{proof}
Multiplying \eqref{eq:overlap} by $X$ gives \eqref{eq:forced}; the norm inequality is submultiplicativity. Let $M=J^\top X$. By Proposition~\ref{prop:backward}, the gradient change for label $y$ is $M(p-e_y)$. If this vanishes for every $y$, all columns of $M$ equal $Mp$. Since $M\mathbf1=J^\top X\mathbf1=0$, their common value is zero. The converse is immediate.
\end{proof}
The all-label statement is stronger than a claim about the particular observed label: a nonzero matrix can annihilate that label's residual. Nevertheless the theorem gives a structural obstruction to treating coordination as ``repair without ever affecting acquisition.'' Whether the forced channel change helps the actual task remains a separate response and value question.

\subsection{Acquisition can enlarge protected repair capacity}
The reverse direction has a constructive characterization. Suppose a lawful acquisition makes $k$ new features available to a \emph{subsequent} regulation action. Their current protected-frame responses are $B\in\R^{m\times k}$ and their obligation-frame responses are $\Psi\in\R^{p\times k}$. Old and new coefficients $X\in\R^{d\times q}$ and $Y\in\R^{k\times q}$ must satisfy
\begin{equation}
 CX+BY=0,\qquad HX+\Psi Y=D.\label{eq:joint-repair}
\end{equation}
We assume both coefficient blocks are actually writable, share the stated output semantics, and have squared cost $\norm{X}_F^2+\norm{Y}_F^2$. A deployed canonical bank need not have this enlarged action interface.

\begin{theorem}[Compensation-adjusted acquisition of repair directions]\label{thm:compensation}
No full-row-rank assumption on $C$ is required. Define
\begin{equation}
\begin{aligned}
 P_C &= I-C^\dagger C, & R_C &= I-CC^\dagger,\\[3pt]
 N &= HP_C, & \Ran Q &= \ker(R_CB),\\[3pt]
 Q^\top Q &= I, & L &= (\Psi-HC^\dagger B)Q,\\[3pt]
 W &= I+Q^\top B^\top(C^\dagger)^\top C^\dagger BQ.
\end{aligned}
\label{eq:compensation-objects}
\end{equation}
The columns of $Q$ form any orthonormal basis of that kernel; a zero-dimensional basis is allowed. Then:

\emph{(i)} The achievable obligation space of \eqref{eq:joint-repair}, column by column, is $\Ran[N\ L]$. Its increase over old protected repair is
\begin{equation}
 \nu_{\rm repair}=\rank[N\ L]-\rank N\ge0.\label{eq:repair-newrank}
\end{equation}
\emph{(ii)} Let $G_+=NN^\top+LW^{-1}L^\top$. A target is feasible exactly when $G_+G_+^\dagger D=D$. For such a feasible target, the minimum squared coefficient cost is
\begin{equation}
 E_+(D)=\tr(D^\top G_+^\dagger D).\label{eq:compcostgeneral}
\end{equation}
Every old-feasible target remains feasible and has $E_+(D)\le E_0(D)$, where $E_0$ is its old minimum cost.

\emph{(iii)} If $N$ has full row rank, put $G_0=NN^\top$ and $A_D=L^\top G_0^{-1}D$. Then
\begin{align}
 E_+(D)&=\tr\!\left[D^\top(G_0+LW^{-1}L^\top)^{-1}D\right],\label{eq:compcost}\\
 E_0(D)-E_+(D)
 &=\tr\!\left[A_D^\top(W+L^\top G_0^{-1}L)^{-1}A_D\right].\label{eq:strictcost}
\end{align}
The improvement in this cost is strict if and only if $A_D\ne0$.
\end{theorem}
\begin{proof}
The current constraint first requires $R_CBY=0$, hence $Y=QV$. All its solutions are
\begin{equation}
 X=-C^\dagger BQV+Z,\qquad CZ=0.\label{eq:compensatingX}
\end{equation}
The obligation becomes $HZ+LV=D$. The two terms in \eqref{eq:compensatingX} are orthogonal, and $Q$ is an isometry, so
\begin{equation}
 \norm{X}_F^2+\norm{Y}_F^2
 =\norm{Z}_F^2+\tr(V^\top WV).\label{eq:whitened-cost}
\end{equation}
Whitening $V$ by $W^{1/2}$ reduces the problem to a minimum-norm linear equation with response operator $[N\ LW^{-1/2}]$. Unrestricted old coordinates can be projected by $P_C$ without changing their response under $N$ and without increasing cost. Therefore its image is $\Ran[N\ L]$, its output Gram matrix is $G_+$, and the minimum-norm solution gives \eqref{eq:compcostgeneral}. Setting $Y=0$ embeds every old-feasible solution, proving nonincrease without assuming invertibility. When $N$ has full row rank, the Woodbury identity gives \eqref{eq:strictcost}. The middle matrix there is positive definite, so the trace is zero exactly when $A_D=0$.
\end{proof}
The same minimum-norm problem is an instance of the inherited protected repair geometry (Theorem~\ref{thm:repair}) with augmented matrices
$\bar C=[C\ B]$ and $\bar H=[H\ \Psi]$. Consequently, the response Gram matrix also has the exact form
\begin{equation}
 \begin{aligned}
 G_+ &= NN^\top+LW^{-1}L^\top\\
     &=\bar H(I-\bar C^\dagger\bar C)\bar H^\top.
 \end{aligned}\label{eq:augmented-compensation}
\end{equation}
To see the equality directly, let $U$ have orthonormal columns spanning $\ker C$. The columns of
\[
 E=\begin{pmatrix}U&-C^\dagger BQ\\0&Q\end{pmatrix}
\]
span $\ker\bar C$, with $E^\top E=\diag(I,W)$ and $\bar HE=[HU\ L]$. Thus $E\diag(I,W^{-1})E^\top=I-\bar C^\dagger\bar C$, which gives \eqref{eq:augmented-compensation}. The underlying least-norm projection is standard. The decomposition into compatible acquisition directions $Q$, the response after protection compensation $L$, and the cost metric $W$ makes explicit how the acquired structure changes repair reachability and expense.

The theorem identifies the new repair capacity created by acquisition: only $L$, the acquired response remaining after compensation on the protected frame, contributes new admissible repair directions. If $B=0$, all new coefficients are compatible with current protection; choosing $Q=I$ gives $W=I$ and $L=\Psi$. If instead $\Psi=HC^\dagger B$ on the allowable new directions, a nonzero raw historical response can contribute no additional protected repair direction.

Acquisition can therefore benefit regulation in two mathematically distinct ways. It can make a previously infeasible repair target attainable by enlarging $\Ran[N\ L]$; or it can strictly lower the coefficient cost of a target that was already attainable, even when $\nu_{\rm repair}=0$. Both effects are exact for the writable interface and metric in \eqref{eq:joint-repair}; an implementation with fewer candidates must be evaluated on its actual action set.

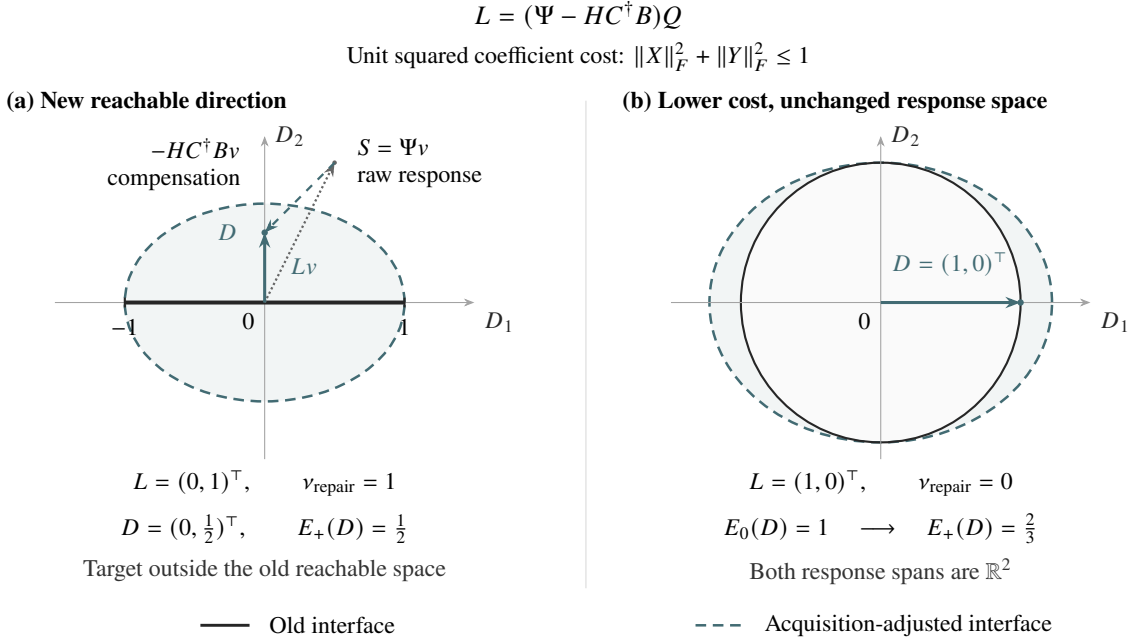
\begin{figure}[H]
  \centering
  \begin{tikzpicture}[x=1cm,y=1cm,
 font=\fontsize{9}{11}\selectfont,
 old/.style={draw=black!85,line width=.8pt},
 joint/.style={draw=pfrteal,line width=.95pt,dash pattern=on 4pt off 2.5pt},
 raw/.style={draw=black!60,densely dotted,line width=.85pt,-{Stealth[length=1.8mm]}},
 comp/.style={draw=pfrteal,dashed,line width=.85pt,-{Stealth[open,length=1.8mm]}},
 effective/.style={draw=pfrteal,line width=1.15pt,-{Stealth[length=2mm]}},
 axis/.style={draw=black!36,line width=.4pt,-{Stealth[length=1.4mm]}}]
\definecolor{pfrteal}{HTML}{426B73}
\path[use as bounding box] (0,.25) rectangle (16,-8.43);
\node[font=\fontsize{10}{12}\selectfont] at (8,-.05)
 {$L=(\Psi-HC^\dagger B)Q$};
\node at (8,-.60) {Unit squared coefficient cost: $\|X\|_F^2+\|Y\|_F^2\leq1$};
\node[anchor=west,font=\fontsize{9}{11}\selectfont\bfseries] at (.30,-1.20)
 {(a) New reachable direction};
\node[anchor=west,font=\fontsize{9}{11}\selectfont\bfseries] at (8.45,-1.20)
 {(b) Lower cost, unchanged response space};
\draw[black!15,line width=.4pt] (8.1,-1.18)--(8.1,-7.63);
\begin{scope}[shift={(3.85,-3.85)},x=1.85cm,y=1.85cm]
 \fill[pfrteal!7] (0,0) ellipse[x radius=1,y radius=.7071067812];
 \draw[joint] (0,0) ellipse[x radius=1,y radius=.7071067812];
 \draw[axis] (-1.50,0)--(1.50,0) node[below right,text=black!85] {$D_1$};
 \draw[axis] (0,-1.12)--(0,1.19) node[right,text=black!85] {$D_2$};
 \draw[old,line width=1.5pt] (-1,0)--(1,0);
 \draw[old] (-1,-.035)--(-1,.035) (1,-.035)--(1,.035);
 \node[below] at (-1,-.045) {$-1$};
 \node[below] at (1,-.045) {$1$};
 \node[below left] at (0,0) {$0$};
 \draw[raw] (0,0)--(.5,1);
 \fill[black!60] (.5,1) circle[radius=.85pt];
 \node[anchor=west,align=left] at (.59,1.00) {$S=\Psi v$\\[-1pt]raw response};
 \draw[comp] (.5,1)--(0,.5);
 \node[anchor=east,align=right] at (-.10,.99)
 {$-HC^\dagger Bv$\\[-1pt]compensation};
 \draw[effective] (0,0)--(0,.5);
 \fill[pfrteal] (0,.5) circle[radius=1.2pt];
 \node[anchor=east,text=pfrteal] at (-.12,.5) {$D$};
 \node[anchor=west,text=pfrteal] at (.11,.26) {$Lv$};
\end{scope}
\begin{scope}[shift={(12.00,-3.85)},x=1.85cm,y=1.85cm]
 \fill[pfrteal!7] (0,0) ellipse[x radius=1.2247448714,y radius=1];
 \fill[black!2] (0,0) circle[radius=1];
 \draw[joint] (0,0) ellipse[x radius=1.2247448714,y radius=1];
 \draw[old] (0,0) circle[radius=1];
 \draw[axis] (-1.5,0)--(1.5,0) node[below right,text=black!85] {$D_1$};
 \draw[axis] (0,-1.12)--(0,1.19) node[right,text=black!85] {$D_2$};
 \node[below left] at (0,0) {$0$};
 \draw[effective] (0,0)--(1,0);
 \fill[pfrteal] (1,0) circle[radius=1.2pt];
 \node[anchor=south east,text=pfrteal] at (.98,.12) {$D=(1,0)^\top$};
\end{scope}
\node at (3.85,-6.24) {$L=(0,1)^\top,\qquad\nu_{\rm repair}=1$};
\node at (12.00,-6.24) {$L=(1,0)^\top,\qquad\nu_{\rm repair}=0$};
\node at (3.85,-6.84) {$D=(0,\tfrac12)^\top,\qquad E_+(D)=\tfrac12$};
\node at (12.00,-6.84) {$E_0(D)=1\quad\longrightarrow\quad E_+(D)=\tfrac23$};
\node[text=black!80] at (3.85,-7.40) {Target outside the old reachable space};
\node[text=black!80] at (12.00,-7.40) {Both response spans are $\mathbb R^2$};
\draw[old,line width=1.1pt] (3.0,-8.12)--(3.65,-8.12);
\node[anchor=west] at (3.78,-8.12) {Old interface};
\begin{scope}[shift={(0.95,0)}]
 \draw[joint] (8.6,-8.12)--(9.25,-8.12);
 \node[anchor=west] at (9.38,-8.12) {Acquisition-adjusted interface};
\end{scope}
\end{tikzpicture}
  \caption{Acquisition-adjusted repair geometry.
  Two independent exact affine examples of Theorem~\ref{thm:compensation}
  are shown in obligation-response coordinates. The solid old set and dashed
  enlarged set contain responses attainable with squared coefficient cost
  at most one. In both examples, $q=1$, $B=Q=1$ and $W=2$.
  (a) Set $C=(1,0)$, $H=\left(\begin{smallmatrix}1&1\\1&0\end{smallmatrix}\right)$
  and $\Psi=(1,2)^\top$. With $v=1/2$, the raw acquired response $S=\Psi v$
  receives the compensation response $-HC^\dagger Bv$, yielding $Lv=D$.
  The target $D=(0,1/2)^\top$ lies outside the old reachable space.
  (b) Set $C=(1,0,0)$, $H=\left(\begin{smallmatrix}1&1&0\\1&0&1\end{smallmatrix}\right)$
  and $\Psi=(2,1)^\top$. Both interfaces span the response plane, yet the
  minimum cost of $D=(1,0)^\top$ falls from $1$ to $2/3$.
  The arrows in (a) express vector addition, not a learning trajectory.}
  \label{fig:acquisition-repair-geometry}
\end{figure}

\subsection{Jointly reachable current responses and infeasibility}
A second characterization asks whether repair, protection, and a desired current learning response can be satisfied within one budget. Include the actual linear actuator image in coordinates $z\in\R^n$. Let $Mz=d$ collect the fixed equality constraints and let $Kz$ be the response of interest.

\begin{theorem}[Joint response ellipsoid and dual obstruction]\label{thm:ellipsoid}
Suppose $Mz=d$ is feasible. Set $z_0=M^\dagger d$, $P=I-M^\dagger M$, and let $\tau\ge\norm{z_0}$ with $\rho^2=\tau^2-\norm{z_0}^2$. The responses obtainable subject to $\norm{z}\le\tau$ are
\begin{equation}
 Kz_0+KP\{u:\norm{u}\le\rho\}.\label{eq:jointellipsoid}
\end{equation}
A target $q$ is attainable precisely when
\begin{equation}
 q-Kz_0\in\Ran(KP),\qquad
 \norm{(KP)^\dagger(q-Kz_0)}\le\rho.\label{eq:jointtarget}
\end{equation}
For $b=Kz_0-q$ and $T=KP$, its minimum residual has the exact dual form
\begin{equation}
 \min_{\norm{u}\le\rho}\norm{b+Tu}
 =\max_{\norm{\lambda}\le1}
       \{\ip{\lambda}{b}-\rho\norm{T^\top\lambda}\}.\label{eq:dualobstruction}
\end{equation}
A strictly positive value certifies current infeasibility at the stated budget.
\end{theorem}
\begin{proof}
Every solution is $z_0+Pu$ with $z_0\perp\Ran P$; Pythagoras gives \eqref{eq:jointellipsoid}. The least-norm preimage of a response difference is $(KP)^\dagger(q-Kz_0)$, giving \eqref{eq:jointtarget}. For any unit-ball $\lambda$ and admissible $u$,
$\norm{b+Tu}\ge\ip{\lambda}{b+Tu}\ge\ip{\lambda}{b}-\rho\norm{T^\top\lambda}$.
For equality, choose a minimizing $u_*$. If its residual is zero, the lower bound zero is attained by $\lambda=0$. Otherwise take $\lambda_*=(b+Tu_*)/\norm{b+Tu_*}$. Convex first-order optimality on the ball implies that $u_*$ minimizes $\ip{T^\top\lambda_*}{u}$ on that ball, whose value is $-\rho\norm{T^\top\lambda_*}$. Substitution attains equality.
\end{proof}
The theorem is a finite-dimensional constrained-response result using standard convex geometry \citep{boyd2004convex}. It is particularly useful for distinguishing ``there is hidden freedom'' from ``the specified repair and acquisition response fit inside the same available action budget.'' A canonical solver that emits only a few candidates must still be tested on those candidates; the full ellipsoid is then a bound or design space, not its automatic menu.

\subsection{Safe reachable directions and finite continuation value}
Joint reachability describes which responses can be realized. Their downstream value is a separate, continuation-dependent question. A local geometric criterion connects the two. Let $\mathcal K$ be a closed convex cone of legally realizable infinitesimal action directions, already incorporating current linear protection and a specified first-order retention condition. At a fixed parent and fixed continuation, assume a differentiable quality $j(a)$ at zero, with $p_j=\nabla j(0)$, and let $\Pi_{\mathcal K}$ denote metric projection. Standard cone geometry gives
\begin{equation}
 \begin{aligned}
 \sup_{a\in\mathcal K,\,\norm{a}\le1}\ip{p_j}{a}
 &=\norm{\Pi_{\mathcal K}p_j},\\[3pt]
 \exists a\in\mathcal K:\ip{p_j}{a}>0
 &\iff \Pi_{\mathcal K}p_j\ne0.
 \end{aligned}\label{eq:safevalue}
\end{equation}
If the ray $ta$, $0\le t\le r$, is actually legal and $\nabla j$ is $M$-Lipschitz along that ray, then for unit $a\in\mathcal K$,
\begin{equation}
 j(ta)-j(0)\ge t\ip{p_j}{a}-\tfrac12Mt^2.
 \label{eq:safefinite}
\end{equation}
For a computed estimate $\widehat p_j$ with $\norm{\widehat p_j-p_j}\le e$, suppose $\delta$ bounds the \emph{contrast discrepancy}:
\[
 |\Delta j_{\rm native}(ta)-[j(ta)-j(0)]|\le\delta.
\]
Then a finite-native lower bound is
\begin{equation}
 \Delta j_{\rm native}(ta)\ge
 t\ip{\widehat p_j}{a}-et-\tfrac12Mt^2-\delta.
 \label{eq:safefinite-native}
\end{equation}
If instead a pointwise error bound $|j_{\rm native}(b)-j(b)|\le\delta_0$ is known at both $b=0$ and $b=ta$, one must use $\delta=2\delta_0$ in \eqref{eq:safefinite-native}.
The equalities follow by orthogonal cone projection and Cauchy--Schwarz; the inequalities follow by integrating the derivative and bounding the gradient error. Positive target value and allowable retention damage require their \emph{separate} bounds. A nonempty protected action space does not make $\Pi_{\mathcal K}p_j$ nonzero, and an unmeasured future covector does not constitute a certified online rule. The criterion characterizes favorable \emph{reachable} directions; the finite bound specifies when that local direction also certifies a favorable realized continuation. It separates the geometry of legal intervention from its future value.

\subsection{A finite learning example of positive complementarity}
A supervised gradient update can make the two operations complementary even when each alone leaves the target response unchanged. The following construction establishes such a case with specified features, loss, and optimizer step.

\begin{theorem}[Complementarity through an actual gradient update]\label{thm:cewitness}
Let $\psi\in\R$ be a trainable representation coordinate, initially zero, and consider two-class logits $W\phi_\psi(x)$ with a head frozen during a single supervised update of $\psi$. On a support example $s$, a historical example $h$, and two query examples $x_\sigma$ with $\sigma\in\{-1,1\}$, define the differentiable features
\begin{equation}
\begin{aligned}
 \phi_\psi(s)       &= (0,\psi,1)^\top,\\[2pt]
 \phi_\psi(h)       &= (1,0,1)^\top,\\[2pt]
 \phi_\psi(x_\sigma)&= (\sigma\psi,0,1)^\top.
\end{aligned}
\label{eq:witness-features}
\end{equation}
The initial head is zero. Regulation retains the historical repair
\begin{equation}
 W_F=\begin{pmatrix}d&0&0\\-d&0&0\end{pmatrix},\qquad d>0,
\end{equation}
and acquisition installs, for the support update only, the temporary component
\begin{equation}
 W_A(q)=\begin{pmatrix}0&bq&0\\0&-bq&0\end{pmatrix},\qquad
 b>0,\quad q\in\{-1,1\}.
\end{equation}
The acquired cue $q$ comes from already available information. It agrees with an unobserved target sign $t\in\{-1,1\}$ with probability $p$, but the future query labels are not available to the action rule. The support and historical labels are class one; for query $x_\sigma$ the correct label is class one exactly when $\sigma t=1$. Perform a single unclipped SGD step of size $\eta>0$ on support cross-entropy, then remove $W_A$ while retaining $W_F$ and the updated $\psi$. Write $\kappa=2d\eta b$.

At installation the support logits are identical in all four combinations, and the historical example obtains the same repaired margin $2d$ whenever regulation is applied. After acquisition removal, the no-action, regulation-only, and acquisition-only policies each give zero query logits and query NLL $\log2$. The joint policy has expected query-NLL improvement
\begin{equation}
 \begin{aligned}
 \mathcal G(p,\kappa)
 &=\frac{(2p-1)\kappa}{2}
   -\log\cosh\!\left(\frac{\kappa}{2}\right),\\[3pt]
 \mathcal G(p,\kappa)>0
 &\iff p>\frac12+\frac{\log\cosh(\kappa/2)}{\kappa}.
 \end{aligned}\label{eq:witness-gain}
\end{equation}
Under a fixed first-class tie rule, the isolated policies have balanced query accuracy $1/2$, while the joint policy has expected query accuracy $p$. In particular, $p=1$ yields strict positive joint NLL improvement over each isolated policy while retaining the specified historical repair exactly.
\end{theorem}
\begin{proof}
The support logits are zero at $\psi=0$, and the regulation head multiplies the zero support coordinate. With the acquisition component, they become $(bq\psi,-bq\psi)$, so the genuine support-loss derivative at zero is $-bq$. The SGD update is therefore $\psi^+=\eta bq$ when acquisition is present and zero otherwise. After the temporary head is removed, the joint query logits equal $(\sigma d\eta bq,-\sigma d\eta bq)$; every other branch yields zero query logits. The correct-class logit margin on either query equals $\kappa qt$, and its cross-entropy is $\log(1+e^{-\kappa qt})$. Using $\Pr(q=t)=p$ and
$\log(1+e^{-z})=\log2+\log\cosh(z/2)-z/2$ gives \eqref{eq:witness-gain}. The two query labels are opposite, hence a zero-logit tie predicts one of two correctly, while joint predictions agree with both precisely when $q=t$. Finally, the historical feature is independent of $\psi$ and annihilates $W_A$, preserving the regulation margin throughout. 
\end{proof}

The example establishes a concrete mechanism: acquisition opens the support-gradient channel, while regulation makes the resulting representation change visible at the queries. Neither action alone yields this response. For a first bias-corrected AdamW step from zero moments with inactive clipping, the same construction replaces $\eta b$ by $\eta b/(b+\epsilon)$ (under the stated denominator convention), so the mechanism is not specific to plain SGD. Positive NLL gain depends on the cue reliability and response magnitude as specified in \eqref{eq:witness-gain}; an equally informed predictor that uses the cue directly may reproduce the query decisions. The construction is an exact learning example, rather than a Qwen measurement or an all-method optimality claim.

\subsection{Propagation through continued learning}
The finite interventions above change the state on which subsequent learning acts. On a common convex chart, suppose $U_t$ is $C^1$ on an open neighborhood of the entire comparison segment between $s_t^0$ and $s_t^1$; assume the compared readout $J$ is likewise $C^1$ on its relevant segments. Let both states follow this same deterministic update. With $\delta_t=s_t^1-s_t^0$,
\begin{equation}
 \delta_{t+1}=\overline J_t\delta_t,\qquad
 \overline J_t=\int_0^1DU_t(s_t^0+u\delta_t)\,du.\label{eq:finitepropagation}
\end{equation}
The line-segment fundamental theorem of calculus then gives the displayed identity, without assuming a constant Jacobian. Consequently $\delta_H=\overline J_{H-1}\cdots\overline J_0\delta_0$. A readout difference has the analogous path integral through $DJ$. In a piecewise-smooth or discrete system, \eqref{eq:finitepropagation} applies only on qualified branches; actual kernel composition remains the general definition.

When a policy recomputes actions after the state changes, it also changes later inputs to the update. For example, a regulation write can change the gradient read by a later FAT generator, and a renewed representation can change a later repair obligation. This is a forward-time feedback path. It is not represented by multiplying the four heterogeneous blocks of \eqref{eq:fourchannels}. Exact execution retains the generator, its information, the optimizer, and the action sequence in the complete state.

\subsection{From local effects to policy value}
Coupling can help, hinder, or leave a particular readout unchanged. For a valid four-policy comparison from the same parent, write $I=J_{11}-J_{10}-J_{01}+J_{00}$, $a=J_{10}-J_{00}$, and $b=J_{01}-J_{00}$. Then
\begin{equation}
 J_{11}>\max\{J_{10},J_{01}\}
 \quad\Longleftrightarrow\quad
 I>-\min\{a,b\}.\label{eq:gain-not-synergy}
\end{equation}
Thus joint advantage over both separate policies does not require positive superadditivity. For instance, $(J_{00},J_{10},J_{01},J_{11})=(0,3,4,5)$ gives joint advantage but $I=-2$. Moreover a table comparing differently coordinated policies is not automatically a valid factorial experiment.

A rigorous connection to performance is available, but it requires actual continuation value rather than a convenient current proxy. Let $\beta$ be a lawful reference policy on the full compared state space, with terminal utility $J_T$ and finite horizon $T$. Here $K_t^a$ and $K_t^\beta$ denote the successor marginals of the relevant kernels. Its values satisfy $V_T^\beta=J_T$ and $V_t^\beta=K_t^\beta V_{t+1}^\beta$. Define
\begin{equation}
 A_t^\beta(s,a)=(K_t^a-K_t^\beta)V_{t+1}^\beta(s).\label{eq:advantage}
\end{equation}
All relevant costs are included in the state/utility, and actions have common time and stopping semantics.

\begin{proposition}[Conditional value of coordinated intervention]\label{prop:policyvalue}
For any lawful policy $\alpha$ with the same initial law and integrable values,
\begin{equation}
 J(\alpha)-J(\beta)=\sum_{t=0}^{T-1}\E_\alpha
       [A_t^\beta(s_t,a_t)].\label{eq:perf-diff}
\end{equation}
If information available before each decision provides a valid uniform lower bound on the true advantage of every selected extra action at every reached state, and at each decision the policy either chooses an action with nonnegative certified advantage or follows the reference decision rule, then $J(\alpha)\ge J(\beta)$. Strict positivity of at least one expected summand gives strict improvement.
\end{proposition}
\begin{proof}
Conditional expectation turns each summand into
$\E_\alpha[V_{t+1}^\beta(s_{t+1})-V_t^\beta(s_t)]$.
Summation telescopes to $\E_\alpha J_T(s_T)-\E V_0^\beta(s_0)$. The latter expectation equals $J(\beta)$ under the common initial law. The sign conclusion follows term by term.
\end{proof}
This is an application of a standard finite-horizon policy-difference argument, not an additional algorithm or a claim that the required lower bound is cheap or known \citep{wang2026intrinsic}. In particular, a current margin, gradient norm, or finite CE improvement is not automatically such a bound. A probabilistic calibration event supports only its stated conditional guarantee unless failure losses are also controlled. The experiments below test useful coordinated specializations without claiming that they implement this sufficient condition for universal policy improvement.

\section{Transformer and language-model specialization}\label{sec:llm}
We now realize protection, acquisition, and their cross-responses in a causal Transformer with writable head and low-rank adaptation interfaces. The implementation connects the preceding general results to actual gradient updates, structural writes, and two different learning rules.

\subsection{A common complete learner and writable interfaces}
Let $h_\theta(x)\in\R^d$ be the last hidden representation at a supervised token position and let $z(x)=Wh_\theta(x)\in\R^V$. The base transformer is frozen in the case study, while LoRA coordinates within $\theta$ are trainable \citep{hu2021lora,qwen2025qwen25}. A committed compact repair bank contributes to the effective head $W$. The complete state also includes the LoRA factors, optimizer moments and clocks, memory, teacher when present, bank, coordination state, and causal stream position.

A captured current feature frame $C$ and historical frame $H$ give exactly the repair equations $CX=0$, $HX=D$ for a head increment $X=\Delta W^\top$. With margin threshold $\mu$, the historical row $i$ has deficit
\begin{equation}
\begin{aligned}
 m_i &= z_{i,y_i}-\max_{v\ne y_i}z_{i,v},\\[4pt]
 d_i &=
 \begin{cases}
   0, & m_i\ge\mu,\\
   \mu+\zeta-m_i, & m_i<\mu.
 \end{cases}
\end{aligned}
\label{eq:deficit}
\end{equation}
Let $S_i=e_{y_i}^\top-\mathbf1^\top/V$ and $D=\diag(d)S$. Adding this centered row target raises every correct-versus-competitor margin by $d_i$. A passing exact target therefore repairs the stated historical deficit while leaving the current frame unchanged.

The physical compact bank does not expose an arbitrary dense $d\times V$ write. It stores a factor $T$ and the fixed label directions $S$, with
\begin{equation}
\begin{aligned}
 N &= HP_C,\\[3pt]
 T &= \diag(d)(NN^\top)^{-1}N,\\[3pt]
 X &= T^\top S.
\end{aligned}
\label{eq:canonicalbank}
\end{equation}
on its qualified full-row-rank domain. The candidate is solved in FP64, stored in FP32, and checked after storage. Failed rank or numerical checks are not replaced by an unregistered pseudoinverse or ridge repair.

If $Q$ is an orthonormal basis for $\Ran S^\top$, every \emph{representable} dense write has $X=ZQ^\top$ and $\norm{X}_F=\norm{Z}_F$. A response equation $AX=Y$ requires $Y(I-QQ^\top)=0$ and $AZ=YQ$. This is an exact restriction to the bank image. The canonical candidate in \eqref{eq:canonicalbank} is narrower still. Thus Theorems~\ref{thm:repair} and~\ref{thm:ellipsoid} apply only after the chosen source interface is imposed; an ideal null-space variation is not automatically executable.

\subsection{LoRA renewal as an acquisition interface}
For one target projection, write
\begin{equation}
 M=B_AA_A=\sum_{j=1}^r b_ja_j^\top.
\end{equation}
The subscript distinguishes these factors from the observation map $B$. The effective projection is $M_0+\gamma M$, with $\gamma=\alpha/r=2$ in the experiments. At a task-arrival event, $X_t$ contains module-input rows from already arrived examples, $D_t$ their module-output loss gradients, and $O_t$ resident historical inputs. For physical slot $j$, let $K_{t,j}$ contain the seven retained input-factor rows and let $P_{t,j}$ project onto $\operatorname{rowspan}[O_t;K_{t,j}]$.

\paragraph{Replay renewal.}
The replay realization forms its initial unit input direction from
\begin{equation}
 G_{t,j}=D_t^\top X_t(I-P_{t,j}).\label{eq:nativebirth}
\end{equation}
A top right singular vector gives the direction in Proposition~\ref{prop:birthdirection}; the factor $\gamma$ does not change this maximizing direction. The implementation scales this direction by the median norm of the retained input factors. It replaces $a_j$, sets $b_j=0$, and checks the resulting finite current and historical changes. The projected signal is used to propose a candidate; only these finite checks determine acceptance. Different accepted newborn directions after the same deletion can therefore share the same immediate effective matrix, as in \eqref{eq:turnover}.

\paragraph{Distillation renewal.}
The LwF realization also accounts for the retained output span. If $Q_{t,j}$ has orthonormal columns spanning the seven retained $b$ columns, its initial input-direction proposal uses
\begin{equation}
 G^D_{t,j}=\gamma(I-Q_{t,j}Q_{t,j}^\top)
 D_{C,t}^\top X_t(I-P_{t,j}),\label{eq:distillbirth}
\end{equation}
where $D_{C,t}$ is the CE component of the current module-output gradient. The proposed direction is further projected outside the historical inputs and \emph{all} parent input-factor rows; a zero residual is rejected. The remaining unit direction is scaled by the replaced parent row's norm. This final direction need not still maximize \eqref{eq:distillbirth} under the enlarged projection.

The candidate is first staged with $b_j=0$. CE and weighted-KL gradients are then recomputed at that staged state, using cached teacher outputs. A finite, generally nonzero output column is constructed along a CE-descending direction orthogonal to the staged KL gradient within the retained-output complement. Acceptance requires a strict parent-relative CE decrease, the specified finite KL-growth bound, and the historical-damage checks. Thus zeroing $b_j$ is an \emph{intermediate step}, not the final LwF write. Appendix~\ref{app:execution} gives the amplitude and selection rules.

In both realizations, only the selected factor row/column and their moment slices are renewed; the scalar optimizer step and the seven other slots are retained. Neither complete replacement is assumed function-neutral relative to the parent. Temporary expansion followed by compression would be a different capacity interface. A general reparameterization of $A_A,B_A$ preserves neither physical-slot identity nor this intervention's optimizer-state changes.

\subsection{Optimizer-native response}
For AdamW, denote the clipped gradient by $g$, the step counter by $n$, and the moments by $m,v$. The ideal scalar-coordinate update is \citep{kingma2015adam,loshchilov2019adamw}
\begin{equation}
\begin{aligned}
 m^+ &= \beta_1m+(1-\beta_1)g,\\[3pt]
 v^+ &= \beta_2v+(1-\beta_2)g^2.
\end{aligned}
\label{eq:adam-moments}
\end{equation}
\begin{equation}
\begin{aligned}
 \theta^+ ={}& (1-\eta\lambda_w)\theta\\[3pt]
 &-\eta\frac{m^+/(1-\beta_1^{n+1})}
                 {\sqrt{v^+/(1-\beta_2^{n+1})}+\epsilon}.
\end{aligned}
\label{eq:adam-parameter}
\end{equation}
For two branches with the same moment hyperparameters and counter, their exact moment differences satisfy
\begin{align}
 \delta m^+&=\beta_1\delta m+(1-\beta_1)\delta g,\\
 \delta v^+&=\beta_2\delta v+(1-\beta_2)
                  (2g^0\delta g+(\delta g)^2).\label{eq:adam-pair}
\end{align}
The parameter response follows by subtracting \eqref{eq:adam-parameter} with each branch's own moments. Equal raw gradients are insufficient when clipping, moments, or clocks differ. The finite head-gradient identity remains an exact real-arithmetic statement at the declared current cut, while the actual optimizer consumes clipped, numerically realized gradients. This distinction is the consumer boundary of Theorem~\ref{thm:consumer}.

\subsection{Replay and output distillation are different consumers}
With available replay \citep{rolnick2019replay}, the ordinary loss in the first realization is
\begin{equation}
 L_R=\tfrac45\overline\CE_{\rm current}
       +\tfrac15\overline\CE_{\rm replay};\label{eq:replayloss}
\end{equation}
without an old-task pool it is current CE alone. Regulation-linked allocation can later replace an eligible replay item using a signal published by an accepted repair. The head write and experience allocation are distinct operational interfaces on the same learner.

In the second realization there is no ordinary replay training. A frozen teacher provides a token distribution $q_T$, and
\begin{equation}
 \begin{aligned}
 L_D(z,y;q_T)&=\CE(z,y)+\lambda T^2\KL(q_T\|p_T),\\[3pt]
 p_T&=\softmax(z/T),\qquad p=\softmax(z).
 \end{aligned}\label{eq:lwfloss}
\end{equation}
This is a token-level output-distillation specialization inspired by learning without forgetting \citep{li2016lwf}, not a claim to reproduce every part of its original protocol. The archived runs use $T=2$ and $\lambda=1$: the multiplier of the raw temperature-two KL is therefore four. 

\begin{proposition}[Exact distillation response]\label{prop:lwf}
For a fixed teacher, the logit gradient and Hessian are
\begin{align}
 r_D&=p-e_y+\lambda T(p_T-q_T),\label{eq:lwfgrad}\\
 H_D&=\diag(p)-pp^\top
       +\lambda\bigl[\diag(p_T)-p_Tp_T^\top\bigr].\label{eq:lwfhess}
\end{align}
For any finite logit change $\delta$, the exact change in loss is
\begin{equation}
 \begin{split}
 \Delta L_D={}&\log\sum_v p_v e^{\delta_v}-\delta_y\\
 &+\lambda T^2\left[\log\sum_v(p_T)_v e^{\delta_v/T}
                         -q_T^\top\delta/T\right].
 \end{split}\label{eq:finite-lwf}
\end{equation}
A protected head write has acquisition-coordinate gradient change $J^\top Xr_D$. At a fixed student, replacing $q_T$ by $q_T'$ changes its parameter gradient by
\begin{equation}
 \Delta\nabla_\theta L_D=-\lambda T J_z^\top(q_T'-q_T),\label{eq:teacherchannel}
\end{equation}
where $J_z=D_\theta z$.
\end{proposition}
\begin{proof}
Use $\nabla_z\log\sum_v e^{z_v/T}=p_T/T$. Differentiating once more gives the temperature-cancelled covariance in \eqref{eq:lwfhess}. For finite $\delta$, subtract the two log-normalizers and the two linear target terms, giving \eqref{eq:finite-lwf}. The protected head formula follows from Proposition~\ref{prop:backward}; the teacher formula follows by subtracting \eqref{eq:lwfgrad} and multiplying by $J_z^\top$.
\end{proof}
At a teacher-matched token, $q_T=p_T$, the current KL gradient is zero but a finite nonconstant logit change has positive KL cost. A rule based only on that zero first-order gradient therefore misses a genuine finite admission cost. This is why the LwF acquisition specialization checks actual finite CE and KL rather than simply copying a replay-gradient heuristic. Teacher refresh is a separate temporal operation: a committed bank can be inherited by the next teacher, but its effect on a new input is measurable only after that input actually arrives.

\subsection{The instantiated cross-coupling}
Three interfaces connect the theory to the executable joint learner. A head repair changes acquisition's current gradient through $J^\top Xr$, even when its protected logits are unchanged. A structural renewal changes captured features and historical deficits, so the next regulation certificate must use its current state, as in \eqref{eq:residualtransport}. A coordinated action rule can use these arrived functional and acquisition readouts to choose among its genuinely available candidates.

In the replay joint policy, the selected FAT rule is combined with regulation-linked replay allocation. In the distillation joint policy, the rule first prefers eligible acquisition candidates that introduce no new historical margin deficits, then candidates that do not increase the existing total deficit, and then the largest already measured finite current CE gain; prescribed fallback and one-event rank cooldown remain part of the rule. The regulation decision is also benefit-conditioned among the canonical write and no-write. These are finite online preferences, not certified lower bounds on future policy value.

Theorem~\ref{thm:compensation} describes a possible \emph{enlarged joint repair interface}, whereas these experiments use a compact canonical bank and finite structural candidates. They do not claim to solve its unrestricted old-plus-new coefficient problem. The theoretical result remains meaningful as a characterization of when acquisition can enlarge repair capacity; its stronger solver-level instantiation is distinct from the cross-gradient and certificate pathways tested here.

\section{Empirical validation of the specializations}\label{sec:evidence}
We examine two consequences of the theory: regulation and acquisition can change each other's current responses, and coordinated policies can improve learning outcomes beyond either separate policy. The experiments instantiate the interfaces in Section~\ref{sec:llm}, rather than the unrestricted joint solver of Theorem~\ref{thm:compensation}.

\subsection{Setting and evaluation}
The experiments use Qwen2.5-32B with rank-eight LoRA and AdamW on a four-task stream derived from MASSIVE 1.1 en-US \citep{qwen2025qwen25,fitzgerald2022massive}. There are 16 intents, with 64 training and eight evaluation examples per intent: 1,024 training examples and a disjoint 128-item evaluation panel. Each task contains four intents and receives 64 ordinary updates, giving 256 updates per run. The panel was repeatedly examined during development.

We examine two training trajectories. The first uses the original random seed and example sequence. The second uses a new seed and a different permutation of examples \emph{within each task}; the four task groups retain the same order. Both trajectories use the same 1,024 training examples and the same 128 evaluation items. Within each trajectory and learning rule, the four policies begin from the same experimental configuration and share the ordinary-update budget, although their states can diverge after intervention. The second trajectory tests a changed training history, not generalization to new data or the separate effects of seed and order.

\emph{Exact} counts an example only when all answer-label tokens and EOS are correctly predicted under teacher forcing. It is not free-generation exact match. Macro NLL averages supervised-token mean loss within each example, then within and across intents. The balanced panel makes it equal to the mean of the 128 example losses, not to a pooled token loss. The original pilot designated NLL as primary and Exact as secondary; the Exact-first presentation here is retrospective.

Online decisions use arrived training examples and resident historical information, not evaluation scores or unseen future-task inputs. The ordinary-update budget is matched within each learning rule. Auxiliary work is not: in both trajectories, baseline, FRT, FAT, and joint LwF use 771, 771, 783, and 803 teacher forwards, respectively. Candidate probes, sensing, and repair also incur costs. Appendix~\ref{app:execution} records these implementation and information-access details.

\subsection{Independent and coordinated policies}
Table~\ref{tab:mainresults} reports the full-policy endpoints. The coordinated policy exceeds both separate structural policies in Exact in every learning-rule--setting comparison. Its net gains over the stronger separate policy are one example for replay under each trajectory, one for distillation under the original trajectory, and two for distillation under the changed trajectory. It also has the lowest macro NLL under both learning rules in the changed trajectory.

\begin{table}[tb]
\centering\small
\caption{Full-policy outcomes on the same 128-item MASSIVE-derived panel. Exact is teacher-forced label-plus-EOS correctness. Bold marks the highest Exact and lowest macro NLL within each learning rule and trajectory. These are complete-policy comparisons, not fixed-action factorial experiments.}\label{tab:mainresults}
\setlength{\tabcolsep}{4.1pt}
\begin{tabular}{@{}llrrrr@{}}\toprule
 & & \multicolumn{2}{c}{Original trajectory} & \multicolumn{2}{c}{Changed trajectory}\\
 \cmidrule(lr){3-4}\cmidrule(l){5-6}
 Learning rule&Policy&Exact&Macro NLL&Exact&Macro NLL\\\midrule
 Replay&Baseline&102/128&0.279321&110/128&0.249136\\
 &FRT&113/128&\textbf{0.200758}&112/128&0.268962\\
 &FAT&114/128&0.216516&110/128&0.263532\\
 &Joint FRT--FAT&\textbf{115/128}&0.222998&\textbf{113/128}&\textbf{0.240527}\\\midrule
 LwF&Baseline&85/128&0.491530&75/128&0.547257\\
 &FRT&93/128&\textbf{0.404870}&92/128&0.450175\\
 &FAT&87/128&0.455100&77/128&0.539343\\
 &Joint FRT--FAT&\textbf{94/128}&0.440946&\textbf{94/128}&\textbf{0.421768}\\\bottomrule
\end{tabular}
\end{table}

The separate mechanisms exhibit different metric trade-offs. In the changed replay trajectory, FRT raises Exact but worsens NLL, while FAT matches baseline Exact with higher NLL. Under the original trajectory, both joint policies lead in Exact but not in NLL; in the changed trajectory, they lead on both readouts.

\FloatBarrier
\subsection{The changed-trajectory test}\label{sec:changed-trajectory}
Under the changed replay trajectory, FRT and joint each commit three repairs; FAT and joint each commit three structural renewals. Under changed distillation, FRT and joint each commit five repairs, while FAT and joint each commit three renewals. The LwF joint policy selects different later physical slots from FAT alone. This confirms a different action path without attributing the terminal gain to a single operation.

For the changed distillation trajectory, three evaluation items are correct only under the joint policy and one only under FRT, giving the net gain of two. Joint minus FRT Exact across the four tasks is $(-1,0,+1,+2)$. The descriptive four-policy Exact contrast is
\begin{equation}
 I_{\mathrm{LwF},\mathrm{changed}}=94-92-77+75=0,\label{eq:lwf-changed-contrast}
\end{equation}
while the macro-NLL contrast is $-0.020493$. Changed-trajectory replay gives $+1$ on Exact and $-0.042830$ on NLL (computed from the unrounded arm-level NLL values; differencing the displayed six-decimal values gives $-0.042831$). Joint advantage does not require positive superadditivity, as \eqref{eq:gain-not-synergy} shows. These full-policy contrasts also do not replace fixed-parent, fixed-action causal comparisons.

The reported endpoints follow the fixed evaluation rule. Retained predictions, terminal summaries and action records support the numerical comparisons; Appendix~\ref{app:verification-scope} states the verification scope.

\subsection{Current cross-responses and counterexamples}
The current-response checks complement the endpoint comparisons. Protected FRT writes change measured acquisition-gradient responses in the replay realization. Under distillation, a separate diagnostic policy supplies five matched-action finite probes of the certified acquisition opportunity before and after regulation. Their changes include positive, zero and negative values; Appendix~\ref{app:opportunity} defines the measured CE envelope, normalization, and comparison direction. Acquisition paths also change later functional frames, repair eligibility, and allocation. The direct same-parent evidence is stronger for the FRT-to-FAT channel; reverse effects are represented by the exact residual identity and the observed later paths.

Several controls limit the policy interpretation. A standard-replay FRT combined with a strong FAT rule need not beat FAT alone. Removing the joint LwF rule's one-event cooldown leaves its Exact vector unchanged but worsens NLL. These results distinguish admissibility from continuation value and show why coordination requires more than executing every available action. Appendix~\ref{app:evidence-boundaries} retains the corresponding controls and their scientific interpretation.

The language-model runs demonstrate the execution of protected repairs and structural renewals, with higher teacher-forced Exact for the coordinated policy than for either separate policy in all four comparisons. Because the trajectories share a previously examined panel and the policies differ in auxiliary work, these results do not establish wider generalization or compute-matched superiority.

\section{Discussion and conclusions}\label{sec:discussion}
\subsection{Repair, acquisition, and learning response}
FRT and FAT distinguish present function from the learning permitted by its realization. A repair can meet its target while leaving a family of learning responses available (Theorem~\ref{thm:authority}). A neutral installation can leave the current function unchanged while altering the next learning response (Theorem~\ref{thm:birth}). These results identify different ways of acting on hidden structure without requiring regulation and acquisition to preserve the same observation.

The coupling theory concerns how the two action families constrain each other. A repair obligation can force a gradient-channel change (Theorem~\ref{thm:forced}). New features contribute to protected repair only through the directions that remain after compensation (Theorem~\ref{thm:compensation}). Their combined responses must fit the joint constraints and budget (Theorem~\ref{thm:ellipsoid}). The supervised construction in Theorem~\ref{thm:cewitness} demonstrates that this interaction can produce a favorable learning response absent from either action alone. The coupling is a property of the shared learning-state dynamics, not simply the coexistence of two training rules.

\subsection{Relation to earlier work}
The finite head repair and continuation-conditioned interaction were established in \citet{wang2026intrinsic}. The present theory extends those results to repair-compatible learning responses, acquired repair capacity, and jointly constrained dynamical effects. Standard pseudoinverse, singular-value, and convex-duality methods provide the mathematical instruments; the contribution is the coupled regulation--acquisition structure obtained from them. The experiments assess realizations of this structure, rather than a benchmark-wide ranking of every protected-update or low-rank method.

At the level of complete successor laws, a method that reproduces the same state-dependent outcomes and costs from the same available information has the same continuation value. The distinction between regulation and acquisition therefore rests on their specified action interfaces and effects on learning, not on labels alone.

\subsection{Limits and further questions}
The affine repair and compensation results are exact on their stated action interfaces. Nonlinear protected lifting requires additional regularity and writable directions, as shown in Appendix~\ref{app:nonlinear}. A finite response dimension need not define a sufficient autonomous learning state. Nor does a finite current-frame certificate protect all future inputs, optimizer memory, or later tasks. Establishing useful low-dimensional closed dynamics and finite nonlinear repair domains remains a separate problem.

The sufficient conditions for positive continuation value also require more than current admissibility. Their future-value covectors or lower bounds may be unavailable or costly to estimate. The experiments instead evaluate concrete policies based on arrived information. Their joint Exact advantage is observed in two Qwen2.5-32B trajectories differing in seed and within-task example order on one reused panel; the task-group order remains fixed. New evaluation items, separate seed and order interventions, other model families, and matched end-to-end budgets would test wider empirical scope.

\Needspace{8\baselineskip}
\subsection{Conclusion}
Regulation and acquisition act on the same learning state but answer different questions: how a specified function can be repaired without violating current protection, and how trainable structure can be formed or renewed for subsequent updates. This paper characterizes how those answers change when the operations are combined. It identifies learning-response freedom remaining after protected repair, conditions under which repair must alter a learning channel, and the compensation-adjusted directions through which acquisition changes protected repair reachability and minimum coefficient cost. Joint-response conditions and a supervised cross-entropy construction show that the interaction can be beneficial even when neither intervention alone changes the target outcome.

In Qwen2.5-32B, protected head repair and fixed-capacity LoRA renewal are executable under replay and output distillation. The coordinated policies achieve higher teacher-forced Exact than either separate policy in all four trajectory--learning-rule comparisons. In these policies, available state readouts and arrived evidence guide the selection of admissible repairs and renewals; ordinary learning then resumes from the modified state. Hidden learning structure is therefore both a source of variation in future learning responses and, within specified information and action constraints, a target of intervention.

\paragraph{Data and computational materials.}
The experiments use a pinned Qwen2.5-32B model and MASSIVE-derived data. The two trajectories reuse the same training and evaluation items. Compact execution records and the original complete-state checkpoints were retained for offline checks, but are not included in the manuscript source package. Appendix~\ref{app:execution} states the execution protocol and the scope of numerical verification.

\par\medskip
\begingroup\fontsize{9.2}{10.9}\selectfont\urlstyle{same}\interlinepenalty=10000
\setlength{\bibsep}{1.1pt}
\bibliographystyle{unsrtnat}\bibliography{references}
\endgroup
\par\medskip\Needspace{8\baselineskip}\appendix
\section{Additional mathematical details and boundary examples}\label{app:math}
\subsection{What an observation must retain for a joint decision}\label{app:information}
Fix a finite candidate universe $\mathcal P$ whose labels have a common decoder, and let $\mathscr K(s)\subseteq\mathcal P$ be the exact current admissibility relation. There exists a map $k$ such that $\mathscr K=k\circ O$ if and only if \eqref{eq:joint-info} holds. Necessity follows because a map has one value for each observation. For sufficiency, assign to an observation the relation of any state in its fiber; the hypothesis makes that assignment independent of the representative.

If exact decoding fails, a selector that sees only $o=O(s)$ and must be valid for every compatible state has to choose in
\begin{equation}
 \mathscr K_{\rm robust}(o)=\bigcap_{s:O(s)=o}\mathscr K(s).
\end{equation}
This intersection is both necessary and sufficient for such a pointwise robust choice. In a finite observation domain, choosing one available label poses no measurable-selection issue; an infinite stochastic decision problem requires separate measurability assumptions. If HOLD lies in every relation, robust feasibility may be trivial, whereas robust \emph{progress} still requires a non-HOLD element. This is an information condition, not a performance guarantee.

For a linear hidden coordinate $x$, observed as $Mx$, a desired response $Lx$ is exactly recoverable by a linear decoder precisely when $\ker M\subseteq\ker L$. The proof is the factorization argument in Theorem~\ref{thm:consumer}. If new scalar linear measurements are allowed, the smallest number needed is
\begin{equation}
 k_{\min}=\rank\begin{bmatrix}M\\L\end{bmatrix}-\rank M.
\end{equation}
Indeed, each measurement can add at most one row-space direction, and a basis of the missing row directions attains this count. This count concerns arbitrary available linear measurements; a restricted physical sensor may not be able to obtain those rows at all.

\subsection{Source-restricted and weighted action coordinates}
The Euclidean statements in the main text assume that the coordinates really parameterize the allowed linear action space. If physical actions have the form $x=Ez$ with injective $E$ and physical squared size $x^\top Hx$, $H\succ0$, then the coordinate metric is $W=E^\top HE\succ0$. Whitening $\tilde z=W^{1/2}z$ converts the constraints to $MEW^{-1/2}\tilde z=d$ and the response to $KEW^{-1/2}\tilde z$. Theorems~\ref{thm:repair} and~\ref{thm:ellipsoid} then apply to this actual image and metric. Using the identity metric without this transformation would change the minimum-cost problem.

A finite canonical candidate set is different from a full linear image. An ellipsoid containing all representable responses can supply an impossibility result when a target lies outside it, but membership does not prove that the canonical rule emits a target-reaching candidate. Likewise, two separately budgeted writes need not obey a combined budget:
\begin{equation}
 \norm{X_F+X_A}_F^2=\norm{X_F}_F^2+
 2\ip{X_F}{X_A}+\norm{X_A}_F^2.
\end{equation}
The cross term cannot be removed merely by labeling the actions regulation and acquisition.

\subsection{A consumer-coordinate example}\label{app:consumer}
Equal feature kernels do not ensure identical updates under coordinatewise adaptive optimization. Consider one scalar feature $q$ represented either by one coefficient or by two coefficients with duplicated features $(q/\sqrt2,q/\sqrt2)$. At zero coefficients and zero moments, both realizations have the same function and feature kernel. Let $g\ne0$ be the gradient of the single coefficient, and suppose both Adam realizations have the same pre-update counter $n$. Define
\begin{equation}
 a_n=\frac{1-\beta_1}{1-\beta_1^{n+1}},\qquad
 b_n=\sqrt{\frac{1-\beta_2}{1-\beta_2^{n+1}}}.
\end{equation}
The one-coordinate update is $-\eta a_ng/(b_n|g|+\epsilon)$. Each duplicated-coordinate gradient is $g/\sqrt2$, so summing their functional contributions gives the ratio
\begin{equation}
 \frac{\text{duplicated functional change}}{\text{single functional change}}
 =\frac{b_n|g|+\epsilon}{b_n|g|/\sqrt2+\epsilon}
 \longrightarrow\sqrt2\quad(\epsilon\to0).
\end{equation}
Plain SGD with matched scalar learning rate instead gives equal functional changes in these two realizations. The example assumes zero moments and an explicit common counter; it does not rotate arbitrary nonzero diagonal Adam second moments as if they were a full covariance. Its purpose is to separate feature-space novelty from optimizer-conditioned learning response, not to claim that duplication improves quality. Parameterization dependence is also an active object of low-rank optimization research \citep{castin2026balanced}.

\subsection{A local nonlinear bridge, not a global finite guarantee}\label{app:nonlinear}
The compensation theorem is exact for affine action responses. A standard implicit-function argument identifies the additional condition needed to retain its local meaning in a nonlinear interface.

\begin{proposition}[Local protected lifting]
Let $F(u,v)\in\R^m$ be a $C^2$ protected observation and $G(u,v)\in\R^p$ a $C^1$ obligation readout near $(0,0)$, where $u\in\R^d$ are old writable coordinates and $v\in\R^k$ are newly available coordinates. Put
$C=D_uF(0,0)$, $B=D_vF(0,0)$, $H=D_uG(0,0)$, and $\Psi=D_vG(0,0)$.
If $C$ has full row rank, choose an orthonormal basis $N_C$ of $\ker C$. There is a local $C^2$ parameterization
\begin{equation}
 u(\xi,v)=N_C\xi+C^\dagger\eta(\xi,v),\qquad
 F(u(\xi,v),v)=F(0,0),
\end{equation}
with $\eta(0,0)=0$, $D_\xi\eta(0,0)=0$, and $D_v\eta(0,0)=-B$. The derivative of its obligation response is
\begin{equation}
 D_{(\xi,v)}G(u(\xi,v),v)\big|_0
 =[HN_C\quad\Psi-HC^\dagger B].\label{eq:ift-effective}
\end{equation}
\end{proposition}
\begin{proof}
Apply the implicit function theorem to
$\widetilde F(\eta,\xi,v)=F(N_C\xi+C^\dagger\eta,v)-F(0,0)$.
Its derivative in $\eta$ is $CC^\dagger=I_m$. Differentiating the resulting identity gives the stated derivatives of $\eta$, since $CN_C=0$. The chain rule then yields \eqref{eq:ift-effective}.
\end{proof}
The compensation-adjusted channel reappears, but only locally and under surjectivity of the old writable protection derivative. The proposition does not prove that a source-bound discrete actuator follows this local parameterization, that an arbitrary finite target lies in its range, or that its path stays inside a fixed resource budget. A singular derivative requires a different constant-rank or higher-order analysis; it cannot be rescued by merely substituting a pseudoinverse into a nonlinear update.

\subsection{Intervention value need not be inferable from current protection}
Take two future environments $\omega\in\{-1,1\}$ that agree on all currently available information, and a feasible action $a\in[-1,1]$ with continuation value $V_\omega(a)=\omega a-a^2$. Any nonzero deterministic action has negative value in at least one environment. For a randomized current action, averaging the two environmental expected values gives $-\E[a^2]\le0$. Therefore no rule using only the shared past can guarantee strictly positive value in both worlds without an additional assumption connecting that past to the future.

This elementary counterexample does not invalidate useful structure-aware policies in a specified learning environment. It identifies the extra role of the task family, consumer, and continuation conditions. Theorems about admissible directions and effective repair capacity do not need such foresight assumptions; unconditional policy-improvement claims do.

\clearpage
\section{Execution and evaluation specification}\label{app:execution}
\subsection{Model, data, and ordinary training}
\begin{longtable}{@{}p{40mm}p{114mm}@{}}
\caption{Execution settings of the tested model and learning rules.}\label{tab:settings}\\
\toprule Object & Specification\\\midrule\endfirsthead
\toprule Object & Specification\\\midrule\endhead
Model & Qwen2.5-32B base; 64 layers; hidden size 5120; vocabulary 152064; BF16 backbone\\
Revision & \texttt{1818d35814b8319459f4bd55ed1ac8709630f003}\\
Adapter & FP32 LoRA, rank 8, alpha 16, dropout 0, both factors trainable; q/v/o/down projections in all 64 layers; 30,408,704 trainable adapter parameters\\
Renewal location & Zero-based layer 32, self-attention \texttt{o\_proj}; eight physical factor slots of width 5120\\
Optimizer & AdamW, learning rate $10^{-4}$, $\beta_1=0.9$, $\beta_2=0.999$, $\epsilon=10^{-8}$, weight decay 0, clipping norm 1, constant scheduler; foreach/fused disabled\\
Recorded software & PyTorch 2.10.0, CUDA wheel 12.8, Transformers 4.57.6, PEFT 0.18.1, Accelerate 1.12.0, NumPy 2.3.5; math SDPA, nonreentrant checkpointing, TF32 disabled\\
Training trajectories & Original seed 33093001; changed seed 33102020 and within-task example permutation; four tasks, 16 intents, 1,024 training and 128 development examples; maximum length 128 tokens\\
Updates & Four current examples per ordinary update; 64 updates per task, 256 total; gradients accumulated over single-example microbatches\\
Replay & After task 0, 16 resident old examples per update; current/replay loss weights $4/5$ and $1/5$; 3,072 historical presentations total\\
Resident memory & Per-intent keyed bottom-$k$ reservoir, $k=8$, capacity 128; only arrived current examples admitted\\
FRT & Consultation after every 32 ordinary updates; current protection from last four current examples; historical representatives from strictly older intents; at most 32 token rows per frame\\
FRT constants & Margin threshold 1; buffer $2^{-10}$; current and historical residual tolerances $2^{-12}$; current CE tolerance $2^{-11}$; increment-norm cap 1; relative rank tolerance $10^{-10}$\\
FAT events & After updates 65, 129, and 193; the first ordinary update of the new task has already occurred\\
Distillation & No ordinary historical replay; $T=2$, $\lambda=1$, loss CE$+4\KL(q_2\|p_2)$; teacher captured after boundary maintenance at updates 64, 128, 192\\
Evaluation & Teacher-forced answer-plus-EOS Exact and macro-intent NLL; fixed 128-item development panel; evaluation is not an online policy input\\\bottomrule
\end{longtable}

The task groups, in order, are
\begin{quote}\small
\texttt{alarm\_query, alarm\_set, audio\_volume\_mute, audio\_volume\_up};\\
\texttt{calendar\_query, calendar\_remove, calendar\_set, cooking\_recipe};\\
\texttt{datetime\_query, email\_query, email\_querycontact, email\_sendemail};\\
\texttt{general\_joke, general\_quirky, iot\_cleaning, iot\_coffee}.
\end{quote}
The learner seed initializes Python, NumPy and both CPU/CUDA PyTorch generators before model loading and adapter creation. The pinned PEFT default initialization is retained (Kaiming-uniform input factors and zero output factors); with \texttt{bias=none} and \texttt{use\_rslora=False}. The adapter tensors are then stored in FP32. The tested device is a single NVIDIA RTX PRO 6000 Blackwell; deterministic algorithms and cuDNN determinism are enabled, benchmarking and reduced-precision BF16 matmul reduction are disabled, and \texttt{CUBLAS\_WORKSPACE\_CONFIG=:4096:8}. These details specify numerical execution rather than an additional intervention.

Each intent has 64 training examples and eight development examples, selected from MASSIVE 1.1 en-US. The exact prompt (including newlines) is
\begin{quote}\small\ttfamily
Identify the intent of the following utterance.\\
Utterance: \{utterance\}\\
Intent:
\end{quote}
followed by one newline. The target string is the intent identifier with underscores replaced by spaces, followed by the tokenizer EOS. For example, \texttt{email\_querycontact} maps to \texttt{email querycontact}; the mapping is a literal string substitution, not a semantic relabeling. Answer-token positions and token IDs are produced by the pinned tokenizer and verified against the original input binding.

The runs used fixed sample bindings and training tapes for the 1,024 training and 128 evaluation records, including source UIDs, split assignments, utterances, labels, the four current UIDs at each ordinary update, and native replay draws. Fixed memory-priority and replay-priority orders determined the pseudorandom choices in each trajectory. These run-specific records were retained for offline checks and are not part of the manuscript source. The frame and allocation rules are described below. Only arrived examples may be consulted online. The four task-group identities and their order remain fixed. In the second trajectory only the within-task example permutation, memory sampling, and seed-dependent random draws change; no new evaluation examples are introduced.

For development item $u$ with $L_u$ supervised positions,
\begin{align}
 \operatorname{Exact}(u)&=\mathbf1\{\arg\max_v z_{u,k,v}=y_{u,k}\text{ for every }k\},\\
 \ell_u&=-L_u^{-1}\sum_{k=1}^{L_u}\log p(y_{u,k}\mid x_u,y_{u,<k}),\\
 \operatorname{MacroNLL}&=\frac1{16}\sum_{g=1}^{16}\frac1{8}
                   \sum_{u:\operatorname{intent}(u)=g}\ell_u.
\end{align}
EOS enters both learning loss and Exact, but is excluded from the selected FRT repair frames. Preservation of those frames therefore cannot be equated with preservation of every Exact decision. The initial experimental comparison designated NLL as primary and Exact as secondary; the present Exact-first display is a retrospective presentation choice.

\subsection{Ordering, state, and transactions}
Before an ordinary update, native replay is drawn from the resident pool of strictly older tasks, then the current arm may redirect one item. The update completes all weighted backwards, clips once, updates optimizer and scheduler, and clears gradients. Its four current examples are then admitted to memory, before maintenance; replay examples are not readmitted. A scheduled FRT consultation then stages the canonical repair against fixed current and historical frames. The stored candidate is checked and either committed or rolled back. A replay signal is published only after its associated repair passes; it can affect subsequent replay positions, not the completed update. At a teacher boundary the resulting student, including its committed bank, is copied into the frozen teacher.

At an acquisition event, candidates are compared from a retained common parent. Temporary factor changes, gradients, optimizer state and RNG are restored before another candidate is evaluated. Only the selected physical action persists. Recorded rejection is not counted as a committed birth. The no-write branch may still advance bookkeeping or incur diagnostic cost according to its declared contract.

\paragraph{Numerical regulation checks.}
The bank allocates 256 rows and eight commit slots; the run additionally checks the schedule-implied bound of 192 rows and six accepted commits. The compact head is accumulated: $W_{\rm eff}^{+}=W_{\rm eff}^{\rm parent}+X^\top$, not overwritten. The current and historical residuals are maximum absolute entries, not Frobenius residuals. Separately for $C$ and $N=HP_C$, the prescribed full-row-rank screen requires $\sigma_{\min}>10^{-10}\sigma_{\max}>0$ and enough columns. A QR/Gram solve in FP64 produces the FP32 stored factors; there is no ridge, pseudoinverse fallback, or cap rescaling. Outward-rounded arithmetic certifies the stored affine residuals and increment norm, and a stacked-frame inverse-residual test certifies the ideal map. Native acceptance also requires identical current argmaxes, historical margins at least one, finite logits, and current FP32 maximum absolute per-row and absolute mean CE changes at most $2^{-11}$. A second append from the restored bank must reproduce features and logits bitwise. Missing old tasks or zero deficits cause no write; unqualified rank, norm, or finite residuals reject the candidate. Failed restoration or a failed implementation check stops execution rather than authorizing a different action.

The FRT implementation's backward-equivalence certificate compares a compact written model with its reference implementation at the \emph{same written state}. It does not state that the learner's pre-write and post-write gradients agree. That latter equality would generally contradict \eqref{eq:backward}. The certificate probes each distinct stored-label contrast $e_y-e_o$, the common nonlabel component $e_o$, and zero, where $o$ is the smallest vocabulary index outside the stored labels. Both the new increment and cumulative bank are checked on the first two historical feature rows. For each probe, its input-gradient error must be at most $10^{-4}\|g_{\rm ref}\|_2+\gamma_k\|a\|_2$, where $a$ sums the absolute FP64 reference terms componentwise, $\gamma_k=ku/(1-ku)$, $u=2^{-24}$ and $k=2r+16$ for $r$ stored rows. Zero reference gradients are permitted. This is a finite, implementation-specific consistency check, not a scientific selector for helpful gradients.

\subsection{Resident representatives and token frames}\label{app:frames}
Each intent retains the first eight \emph{arrived} training UIDs in its fixed memory-priority order. At a consultation, intents from strictly older tasks are visited in the task/intent order listed above; the first two resident UIDs in lexicographic order represent each intent. This is not selection by loss, margin, or numerical source-example ID. A missing required pair is a contract failure. No old-task representative is used in task zero.

For FRT, the current examples are the four just trained, in batch order. The historical examples are the representatives just defined. In each frame separately, visit the first non-EOS answer prediction position of every example, then the second of every example, and so on. Keep the first occurrence of each pair (complete causal input-token prefix through that prediction position, target token ID); stop at 32 rows. This round-robin and deduplication rule, not a global sort of all token positions, defines $C$ and $H$. The resulting frame rows were recorded for both trajectories.

At FAT events after updates 65, 129, and 193, use the same four current examples and the same representative rule, but take \emph{only the final non-EOS answer prediction position} of each example. There are four current gate rows and respectively 8, 16, and 24 historical gate rows; there is no additional round-robin deduplication in these FAT frames. Their layer-input activations form the historical protection rows $O_t$. The location is the module with suffix \texttt{layers.32.self\_attn.o\_proj}; its adapter matrices are $8\times5120$ and $5120\times8$ in FP32. All frames are recaptured at the reached parent, including its effective repair bank.

\paragraph{Proposal signals are consumer-specific.}
Replay computes the current example's \emph{whole answer-plus-EOS mean CE}, differentiates it with respect to the selected module's sequence output, and retains the derivative and input activation only at that example's final non-EOS answer prediction position. The four rows are concatenated, with no additional factor $1/4$ in this proposal readout. Thus the derivative includes downstream causal effects of that position; it is not a single-token CE derivative or the ordinary mixed replay gradient.

Distillation instead retains \emph{all sequence positions} of each current example, including prompt positions. It differentiates the same answer-plus-EOS mean CE and the weighted $4\KL$ separately, and divides each example's module derivatives by four before concatenation. Prompt-position derivatives can be nonzero through causal attention. The frozen teacher is evaluated before student hooks are installed. Historical protection and finite gates still use the single-position rows above. Proposal-gradient calls do not advance the optimizer or accumulate parameter gradients, and restore training mode and random-generator states.

\subsection{Regulation-linked replay}\label{app:allocation}
Let $s$ be the last accepted repair's update number. For each historical intent represented in that repair frame, publish a Boolean $r_g$: it is true exactly when at least one of that intent's \emph{pre-repair} row margins was below one. The signal is active at updates $s<t\le s+32$. A new accepted repair replaces the signal and resets its counter; rejection leaves the previous signal unchanged, subject to expiration.

At update $t$, let $N_t$ be the ordered native 16-example replay batch and $M_t$ the pre-admission resident old-task pool. A donor position is eligible exactly when its intent was represented in the signal, has $r_g=0$, and has a same-source-task recipient in $M_t\setminus N_t$ whose intent has $r_g=1$. Choose the eligible position first in the fixed donor-priority order, then the eligible recipient first in that source task's recipient-priority order. Each order is keyed by trajectory, publication update and current update. The exact priority orders were recorded in offline ordinal tables. Ties in the original priorities are resolved by position for donors and lexicographic UID for recipients.

Replace at most one position. With no active signal, no legal exchange, no native replay, or 32 replacements already consumed by that signal, leave the native batch unchanged. ``Unused'' means absent from $N_t$, not absent from all earlier batches in the window. Baseline and acquisition-only replay always keep $N_t$; regulation and joint use their own reached-state signals. This preserves 16 distinct replay UIDs, the source task at every position, and the total loss weight $1/5$. The whole-run count depends on the accepted signal windows. The resulting replay-allocation paths for the changed trajectory were retained for offline checks.

\subsection{Acquisition rules for the two consumers}
At an acquisition event at ordinary update $t\in\{65,129,193\}$, both consumers enumerate at most four candidate physical slots from \texttt{np.random.default\_rng(680016+t).permutation(8)[:4]} (NumPy 2.3.5); the eight slot indices are zero-based. This candidate-order seed is distinct from each trajectory's learner seed. Every trial restores its candidate-specific temporary state unless that trial is selected.
For the three events this gives the ordered slot lists $(5,3,4,6)$, $(7,3,5,2)$, and $(2,5,7,0)$, respectively. Replay-only acquisition and joint replay use the identical renewal generator and first-passing selection rule; their reached parameters and replay histories can differ.

\paragraph{Replay direction and admission.}
Visit the slots in the stated order. For slot $j$, form a row-space basis of $[O_t;K_{t,j}]$ by retaining singular values greater than $\max(10^{-8}\sigma_1,10^{-14})$. Put $R=X_t(I-P_{t,j})$ and use its thin SVD $R=U\Sigma V^\top$. A leading right singular vector of $D_t^\top U\Sigma$, multiplied by $V^\top$, gives the unit input direction in \eqref{eq:nativebirth}. Choose its sign so that its first maximum-absolute-value coordinate is nonnegative, then scale by the median of the seven retained input-row norms. Store the row in FP32 and set the selected output column to zero. Replay does \emph{not} use the symmetric CE-amplitude probes of the distillation rule.

A proposal is unavailable if the largest singular value of $R$ is at most $10^{-10}$, that of $D_t^\top U\Sigma$ is at most $10^{-12}$, or the median input-row norm is at most $10^{-12}$. After FP32 storage, require current-response RMS at least $10^{-6}$ and $\|D_t^\top X_ta_j^{\rm new}\|\ge10^{-12}$. The maximum absolute responses on $O_t$ and the retained rows must each be at most
\[
 10^{-5}\max\!\left\{1,\ r_A\max\!\left(\max_i\|O_{t,i}\|,\max_i\|K_{t,j,i}\|\right)\right\},
\]
where $r_A$ is the median norm. Nonfinite inputs or an invalid/missing required frame do not trigger an alternative estimator.

For the merged four-current-plus-historical token frame, let $z^0,z^1$ be the parent and trial logits, $m_i^b=z^b_{i,y_i}-\max_{v\ne y_i}z^b_{i,v}$, $c_i^b=\CE(z_i^b,y_i)$, and $h_i^b=\mathbf1\{\arg\max_vz^b_{i,v}=y_i\}$. The finite damage gate is
\begin{equation}
 \begin{aligned}
  \max_{i,v}|z^1_{i,v}-z^0_{i,v}|&\le1,
  &\max_i(m_i^0-m_i^1)_+&\le1,\\
  \max_i(c_i^1-c_i^0)&\le0.25,
  &\sum_i h_i^0(1-h_i^1)&\le1.
 \end{aligned}\label{eq:renewal-finite-gate}
\end{equation}
All logits must be finite. These are maxima and a count over the \emph{combined token rows}, not over complete examples or separately allowed budgets for current and historical data. The finite gate computes margins, token CE and logit differences in CPU FP32 from the stored logits. Argmax ties select the first vocabulary index. No strict CE decrease is required by replay renewal.

Accept the first available proposal passing \eqref{eq:renewal-finite-gate}; otherwise restore the parent and try the next slot, ending in HOLD if all fail. On commit, reset the selected row/column slices of the first and second moments (and the maximum second moment if present). Keep the optimizer's scalar age and all seven other slots unchanged. Rank and parameter count remain fixed.

The distillation rule is a distinct, finite-capacity renewal with a current CE/KL admission test. After the first ordinary update of each new task, it captures four arrived training examples and the existing teacher. For the same four keyed candidate slots, it forms a projected current CE direction outside historical input and retained-factor spans; the slot's original input-factor norm sets the new input-factor norm. It stages the proposed input row with the selected output column zero. Let $g_C,g_K$ be the \emph{actual} staged-state gradients of the four-example mean CE and weighted teacher KL with respect to that output column. Project both outside the other seven output columns to obtain $u_C,u_K$. The normalized CE-descent, first-order KL-neutral direction is
\begin{equation}
 w=-\frac{u_C-(\langle u_C,u_K\rangle/\|u_K\|^2)u_K}{\|u_C-(\langle u_C,u_K\rangle/\|u_K\|^2)u_K\|},\label{eq:distil-direction}
\end{equation}
when $\|u_K\|^2>10^{-24}$; otherwise set the numerator to $u_C$. Reject if its norm is at most $10^{-12}$, or if the resulting CE directional derivative is nonnegative.

\paragraph{Finite output-column amplitude.}
Let $r_B=\|b_j^{\rm old}\|$, $v_j$ be the current output-factor AdamW second-moment column, $\beta_2$ its decay, $n$ the optimizer counter, and $\eta,\epsilon$ the optimizer's current learning rate and stabilizer. Let $P_\perp$ project outside the other output-factor columns. The initial symmetric probe radius is
\begin{equation}
 \begin{aligned}
 \widetilde g_C&=P_\perp\!\left[g_C\oslash\left(\sqrt{v_j/(1-\beta_2^n)}+\epsilon\right)\right],\\
 h_A&=\eta\left|\langle\widetilde g_C,w\rangle\right|,\qquad
 h=\sqrt{r_B\min(r_B,h_A)}.
 \end{aligned}\label{eq:finite-born-radius}
\end{equation}
The method measures mean CE at both $+hw$ and $-hw$ around the staged zero-output state. If both coincide numerically with the zero-column CE, it updates $h\leftarrow\sqrt{hr_B}$, for at most three symmetric probe pairs or until $h\ge r_B$. If all probes are unresolved, the candidate is rejected. The counter must be positive, the stored moments finite and nonnegative, and both $r_B$ and $h_A$ positive; otherwise no candidate amplitude is available. From the final pair let
\begin{equation}
 \begin{aligned}
 \ell&=-\langle g_C,w\rangle>0,\qquad
 k_h=\frac{\CE(+hw)+\CE(-hw)-2\CE(0)}{h^2},\\
 \alpha_B&=\frac{\ell}{\max(k_h,0)+\ell/h},\qquad
 b_j^{\rm new}=\operatorname{FP32}(\alpha_B w).
 \end{aligned}\label{eq:finite-born-amplitude}
\end{equation}
Write $B_C=-\langle g_C,b_j^{\rm new}\rangle>0$. The same arrived four-example frame and cached teacher logits define the parent-relative full finite loss changes. All required row/column bases use the singular-value threshold $\max(10^{-8}\sigma_1,10^{-14})$; a nonfinite or at-most-$10^{-12}$ projected direction is unavailable. Singular-vector/eigenvector ties follow the specified FP64 linear-algebra backend; after normalization the first maximum-absolute coordinate fixes the sign. A candidate qualifies only if the repeated CE/LwF readouts agree exactly (FP32 per-example losses averaged in FP64, with no comparison tolerance), $\Delta\CE<0$, $\Delta(4\KL)\le B_C$, and the fixed finite historical/current damage gate passes (at most one previously correct evaluated row may turn incorrect, maximum absolute-logit change $1$, margin decrease $1$, and per-row CE increase $0.25$). This finite test, not first-order orthogonality alone, decides acceptance.

\paragraph{Selecting a structural renewal.}
For acquisition alone, sort the qualifying physical slots by descending projected CE singular value, then ascending finite $\Delta\CE$, then the keyed proposal order. Exclude the slot most recently \emph{successfully} renewed; select the first remaining candidate or HOLD. An intervening HOLD does not clear this one-event cooldown. For the joint rule, begin with the same qualified and noncooled candidates and prefer, in order: candidates that create no new historical FRT margin deficit; among them candidates whose \emph{total} historical FRT deficit does not increase; among that subset the largest observed finite current CE reduction, with the preceding ranking as tie-break. At any stage with an empty preferred subset, use the preceding stage's choice; the final fallback is the original acquisition choice, otherwise HOLD. These are current-state preferences, not future-value estimates.

\paragraph{Selecting a regulation write under distillation.}
The joint FRT consultation compares the native certified repair with HOLD. Let $B_H=\overline\CE_H(s)-\overline\CE_H(f(s))$ be its actual mean \emph{token-row} historical CE gain on the qualified FRT frame (FP32 bank correction followed by FP64 CE reduction), and let $\Delta_C$ and $\Delta_K$ be the differences of the four-example mean CE and raw temperature-two KL (repair minus HOLD) under the same teacher. The gate computes $\Delta_D=\Delta_C+4\Delta_K$ from these separately reduced scalars; it does not subtract two already-rounded combined-loss scalars. Commit the native repair precisely when $B_H\ge0$, $\Delta_C\le B_H$, and $\Delta_D\le B_H$; otherwise HOLD. This is a declared 1:1 current-loss trade-off, not a forecast of terminal Exact. All compared candidates are evaluated from the same restored parent; only an accepted physical state change persists.

\subsection{Resources and information access}
The distillation learner has no ordinary replay training, but FRT and FAT can still read resident historical representatives for current sensing and qualification. It is therefore not a memory-free full-system comparison. Teacher operations, finite candidate probes, frame acquisition, repair solves and storage are additional work. The last-stage joint ranking reuses already measured candidate readouts; this does not make the candidate generation or the whole policy free. No matched end-to-end FLOP or wall-clock superiority claim is made.

\subsection{Changed training trajectory and computational accounting}\label{app:changed-protocol}
The second trajectory uses random seed $33102020$, rather than the original $33093001$, and a different permutation of the training examples within each of the four tasks. The ordered task groups remain unchanged. Both runs use the same 1,024 training examples and 128 previously inspected evaluation examples, with 64 ordinary updates per task. The reordering and seed jointly change the reached learning state; this design does not identify their separate effects. Each policy is evaluated using the same teacher-forced endpoint definition within its trajectory.

\begin{table}[ht]
\centering\small
\caption{Committed structural operations in the changed training trajectory. Repair consultations may decline to write. Replay redirections count individual examples substituted during subsequent batches.}\label{tab:changed-actions}
\begin{tabular}{@{}llrrr@{}}\toprule
Learning rule&Policy&FRT commits&FAT commits&Replay redirections\\\midrule
Replay&Baseline&0&0&0\\
&Regulation&3&0&95\\
&Acquisition&0&3&0\\
&Joint&3&3&96\\\midrule
Distillation&Baseline&0&0&0\\
&Regulation&5&0&0\\
&Acquisition&0&3&0\\
&Joint&5&3&0\\\bottomrule
\end{tabular}
\end{table}

In replay, each accepted repair can publish a 32-update allocation signal; at most one item is redirected per subsequent update. The recorded regulation-only run has 32, 32, and 31 replacements in its three active windows, and the joint run has 32 in each. These allocations were recomputed from the retained controller records. Compact records allow replay-policy and numerical endpoint reconstruction; this does not constitute independent full-tensor re-execution of each model update.

\begin{table}[ht]
\centering\small
\caption{Teacher forward calls in distillation. Both training trajectories have the same count for each policy. Extra calls are relative to the corresponding baseline.}\label{tab:teacherbudget}
\begin{tabular}{@{}lrr@{}}\toprule
Policy&Teacher forwards per trajectory&Extra calls\\\midrule
Baseline&771&0\\
Regulation&771&0\\
Acquisition&783&12\\
Joint&803&32\\\bottomrule
\end{tabular}
\end{table}

Here one Teacher forward means one teacher-model evaluation of one utterance example (including its supervised answer-token positions). The common count $771=3\times64\times4+3$ comprises 768 ordinary teacher calls during tasks 1--3 and one snapshot-consistency check at each of three task boundaries. Acquisition adds four such calls at each of three structure events ($12$), and the joint regulation-selection diagnostic adds four calls at each of five qualified repair consultations ($20$); thus the acquisition and joint totals are $783$ and $803$.

\Needspace{3\baselineskip}
Teacher forward count does not exhaust extra FLOPs, memory use, or wall-clock costs. Candidate diagnostics, finite CE/KL checks, regulation solves, and state inspection also consume resources. The policies match ordinary-update counts but are not claimed to match full computational expenditure.

\section{Additional empirical controls and scope}\label{app:evidence-boundaries}

\subsection{Comparability of the main policies}
The replay baseline in Table~\ref{tab:mainresults} contains neither structural repair nor acquisition. The regulation-only policy uses both a qualified repair and regulation-linked replay allocation, whereas the acquisition-only policy retains ordinary replay allocation. The successful coordinated replay policy also uses regulation-linked allocation. Thus the policy table compares complete realizations; it is not a fixed-parent factorial isolation of structural repair from replay-content allocation. The distillation policies likewise use separate, fully executed acquisition and joint selection rules.

\subsection{Coordination alternatives}
The following original-seed controls change a stated component of an otherwise fixed policy. In replay, keep the direct renewal, canonical repairs and training tape, but disable all regulation-linked redirections. This ordinary-allocation joint policy reaches 112/128 Exact, versus 114/128 for acquisition alone. The successful joint row in Table~\ref{tab:mainresults} instead includes the allocation rule in Appendix~\ref{app:allocation}. This control distinguishes those complete coordination rules; it does not hold their later learning states fixed.

For distillation, retain the same renewal candidates, finite gates and joint preferences but remove the exclusion of the most recently renewed slot. This no-cooldown variant preserves all 128 binary Exact decisions and the count 94/128, but changes macro NLL from 0.440946 to 0.465207. Thus an unchanged Exact vector need not mean an unchanged predictive distribution. These are development controls on the same reused panel, not additional independent training settings.

\subsection{Finite cross-responses and learning realization}\label{app:opportunity}
The five probes below come from a \emph{separate diagnostic policy} on the original seed and training tape, not from the joint-policy row of Table~\ref{tab:mainresults}. Its ordinary learner, teacher boundaries, frames and FAT-only renewal rule are those of Appendix~\ref{app:execution}. It retains the one-event slot cooldown. Its FRT rule first requires the native repair certificate and the current-loss trade-off in Appendix~\ref{app:execution}, then adds the opportunity test described below. This recurrence identifies the reached parents; equal step numbers alone do not.

\paragraph{A fixed physical proposal from the pre-repair state.}
At a qualified FRT time $t$, freeze the complete pre-write state, including its bank, factors, moments, stream, random state, prior-slot record and teacher. Use the same current four examples and historical representatives as the acquisition frame in Appendix~\ref{app:frames}. Teacher snapshots are at steps 64, 128, 128, 192 and 192 for the five rows, respectively. Enumerate the first four slots from the same keyed permutation as for renewal, excluding the previously accepted FAT slot. Capture the full-token parent gradients $D_C,D_K$, input rows $X$, and historical rows $O$ before appending the FRT bank.

For each available slot, use \eqref{eq:distillbirth} and the subsequent projection outside $O$ and all eight parent input rows to obtain a unit direction $v$. Put $r_A=\|a_j\|$, $a'_j=\operatorname{FP32}(r_Av)$, and form the \emph{parent-state} output gradients
\begin{equation}
 g_C=\gamma r_A D_C^\top Xv,\qquad g_K=\gamma r_A D_K^\top Xv.
 \label{eq:diagnostic-parent-gradients}
\end{equation}
Project outside the other seven output columns and use \eqref{eq:distil-direction} to form $w$. The moment radius $h$ is exactly \eqref{eq:finite-born-radius}, with these parent gradients and moments. Set $b'_j=\operatorname{FP32}(hw)$ and $B_j=-\langle g_C,b'_j\rangle$. Reject an unavailable, nonfinite or nonpositive-radius proposal, or $B_j\le0$. Unlike the scheduled LwF renewal, this diagnostic neither recomputes the gradient at a staged zero column nor performs the symmetric amplitude probes. Among the remaining proposals, retain only the largest $B_j$, resolving ties by enumeration order; the set is empty if none remains.

Freeze that pair $(a'_j,b'_j)$ and its parent budget $B_j$. Evaluate the identical physical write both before and after the bank-only FRT action. In each branch require repeatable finite losses, parent-relative $\Delta\CE<0$, $\Delta(4\KL)\le B_j$, and the combined token-row gate \eqref{eq:renewal-finite-gate}. No proposal is regenerated after FRT. All trial factors and transients are restored; no shadow candidate, optimizer step, or moment reset persists. Within each pair only the regulation bank differs, while teacher and all arrived examples are held fixed.

For the four currently arrived examples $C_t$, let $L_C(s)$ be the mean of their per-example mean answer-plus-EOS token cross-entropies (nats per example). With the same fixed physical proposal set $\mathcal P_t$ and its state-dependent certified subset $\mathcal P_t^{\rm cert}(s)$, define
\begin{equation}
 \begin{aligned}
 Q_C(s;\mathcal P_t)&=L_C(s)-\min\bigl(\{L_C(s)\}
    \cup\{L_C(A_a(s)):a\in\mathcal P_t^{\rm cert}(s)\}\bigr),\\
 \Delta_F Q_C&=Q_C(f(s);\mathcal P_t)-Q_C(s;\mathcal P_t).
 \end{aligned}\label{eq:current-opportunity}
\end{equation}
The baseline HOLD is always included. Hence $Q_C=0$ when the candidate fails the native finite qualification; the certified subsets may differ before and after FRT even though the physical proposals do not. Positive $\Delta_FQ_C$ means regulation makes an immediate certified acquisition opportunity larger; negative means smaller.

\begin{table}[ht]\centering\small
\caption{Matched-physical-proposal finite acquisition opportunity (nats per current example). All entries use arrived training samples and the same teacher within each pair.}\label{tab:current-opportunity}
\begin{tabular}{@{}rrrr@{}}\toprule
Ordinary update&Before FRT&After FRT&$\Delta_FQ_C$\\\midrule
96&0.016320877&0.016660973&$+0.000340097$\\
160&0.009440793&0.010511544&$+0.001070751$\\
192&0&0&$0$\\
224&0.014273338&0.014275268&$+0.000001930$\\
256&0.004792862&0.004691567&$-0.000101294$\\\bottomrule
\end{tabular}\end{table}

At update 192 the shortlisted proposal is uncertified in both branches; the displayed zeros do not show absence of a raw gradient. The tiny positive difference at update 224 is a finite recorded change, not by itself a stable effect. These paired finite checks establish response dependence at the observed states, without identifying an unmeasured future policy value. Regulation also changes current acquisition-gradient responses in the replay realization; later decision rules can use such readouts to select different physical slots.

\paragraph{How these parent states are reached.}
After the probe, let $\Delta_D^{\rm probe}$ be the difference of the recorded four-example mean LwF loss scalars, after minus before FRT. Retain the native write only when $B_H\ge0$, $\Delta_C-\Delta_FQ_C\le B_H$, and $\Delta_D^{\rm probe}\le B_H$, in addition to the preceding native and current-loss gates; otherwise restore HOLD. The combined-loss subtraction here and the separate CE/KL recombination in Appendix~\ref{app:execution} are distinct floating-point readouts of the same real-arithmetic contrast. Ordinary training then continues. On this diagnostic trajectory, FAT commits at updates 65, 129, 193 to slots 6, 7, 0, while accepted FRT writes occur at 96, 160, 192, 224, 256. The shortlisted physical slots at those five probes are 5, 3, 4, 4, 3. Retained diagnostic-anchor records contain the current/historical UIDs, teacher ages, prior-slot state and unrounded scalar readouts. The diagnostic uses 823 teacher-example forwards: the 803-call schedule plus four additional calls at each of the five probe events. These costs are separate from Table~\ref{tab:teacherbudget}.

\Needspace{8\baselineskip}
\subsection{Descriptive four-policy contrasts}
For a common scalar endpoint readout, define the arithmetic comparison
\begin{equation}
 I=Y_{\mathrm{joint}}-Y_{\mathrm{regulation}}-Y_{\mathrm{acquisition}}+Y_{\mathrm{baseline}}.
\end{equation}
A joint policy can outperform both separate policies even when this descriptive Exact contrast is zero or negative. For NLL, lower values are preferable. The complete-policy contrasts are summarized below.
\begin{table}[ht]
\centering\small
\caption{Readout-specific complete-policy contrasts. The policies may follow different later states and actions, so these values are not fixed-action causal factorial effects.}\label{tab:descriptive-cross}
\begin{tabular}{@{}llrr@{}}\toprule
Trajectory&Learning rule&Exact contrast (count)&Macro-NLL contrast\\\midrule
Original&Replay&$-10$&$+0.085045$\\
Original&Distillation&$-1$&$+0.072506$\\
Changed&Replay&$+1$&$-0.042830$\\
Changed&Distillation&$0$&$-0.020493$\\\bottomrule
\end{tabular}
\end{table}
\par\FloatBarrier
For the original setting, the NLL contrasts use the six-decimal endpoints in Table~\ref{tab:mainresults}. For the changed setting, they use the retained unrounded values; differencing the displayed endpoints can change the last reported digit. A mixed contrast identifies a fixed-action interaction only with matched parents and treatments, not merely from four terminal policy rows.

\subsection{Verification scope}\label{app:verification-scope}
The reported endpoints are supported by retained predictions, terminal summaries and compact action records. For the second trajectory, all eight endpoint scores were recomputed from per-item records; those records and the executed replay allocations were retained for offline verification but are not supplied with the manuscript source. Structural-factor snapshots and allocation records were also checked against the specified rules. The original-setting table additionally uses retained terminal summaries where complete per-item records were not available; its displayed precision is unchanged. These are source- and record-consistency checks; they are not additional independent model repetitions, full checkpoint tensor audits, or proofs of statistical generalization. Both trajectories reuse the same evaluation items, which were inspected during development. A new-data comparison, isolated changes of seed and within-task example order, and fully matched computational budgets would be distinct empirical questions. The original experimental comparison prioritized NLL; foregrounding Exact in this manuscript does not retroactively alter that protocol.

\end{document}